\documentclass{article}

\usepackage{microtype}
\usepackage{graphicx}
\usepackage{subfig}
\usepackage{booktabs} 
\usepackage{amsmath,amssymb,amsfonts}
\usepackage{algorithm,algorithmic}

\usepackage{enumitem}
\usepackage{multirow}
\usepackage{amsthm}

\usepackage{hyperref}

\usepackage[accepted]{mlsys2023}

\newcommand{\idlow}[1]{\mathord{\mathcode`\-="702D\it #1\mathcode`\-="2200}}
\newcommand{\id}[1]{\ensuremath{\idlow{#1}}}

\newcommand{\projectName}{L-GreCo}
\renewcommand{\paragraph}[1]{\noindent\vspace{0.2em}\textbf{\emph{#1}}}

\usepackage{array}
\newcolumntype{P}[1]{>{\centering\arraybackslash}p{#1}}
\newcolumntype{M}[1]{>{\centering\arraybackslash}m{#1}}

\mlsystitlerunning{\projectName: Layerwise-Adaptive Gradient Compression}

\begin{document}

\twocolumn[
\mlsystitle{\projectName: Layerwise-Adaptive Gradient Compression for Efficient and Accurate Deep Learning}



\mlsyssetsymbol{equal}{*}

\begin{mlsysauthorlist}
\mlsysauthor{Mohammadreza Alimohammadi}{ist,equal}
\mlsysauthor{Ilia Markov}{ist,equal}
\mlsysauthor{Elias Frantar}{ist}
\mlsysauthor{Dan Alistarh}{ist,neural_magic}
\end{mlsysauthorlist}

\mlsysaffiliation{ist}{Institute of Science and Technology Austria}
\mlsysaffiliation{neural_magic}{NeuralMagic}

\mlsyscorrespondingauthor{Ilia Markov}{ilia.markov@ist.ac.at}

\mlsyskeywords{Distributed Training, Gradient compression}

\vskip 0.3in

\begin{abstract}
Data-parallel distributed training of deep neural networks (DNN) has gained very widespread adoption,  but can still experience communication bottlenecks. 
To address this issue, entire families of compression mechanisms have been developed, including quantization, sparsification, and low-rank approximation, some of which are seeing significant practical adoption.  
Despite this progress, almost all known compression schemes apply compression uniformly across DNN layers, although layers are heterogeneous in terms of parameter count and their impact on model accuracy. 
In this work, we provide a general framework for adapting the degree of compression across the model's layers dynamically during training, improving the 
overall compression, while leading to substantial speedups, without sacrificing accuracy.  
Our framework, called \texttt{\projectName}, is based on an adaptive algorithm, which automatically picks the optimal compression parameters for model layers guaranteeing the best compression ratio while satisfying an error constraint. Extensive experiments over image classification and language modeling tasks shows that \texttt{\projectName} is effective across \emph{all existing families of compression methods}, and achieves up to 2.5$\times$ training speedup and up to 5$\times$ compression improvement over efficient implementations of existing approaches, while recovering full accuracy. Moreover, \texttt{\projectName} is complementary to existing adaptive algorithms, improving their compression ratio by 50\% and practical throughput by 66\%. An anonymized implementation is available at \url{https://github.com/LGrCo/L-GreCo}. 
 
\end{abstract}
]



\printAffiliationsAndNotice{\mlsysEqualContribution} 

\section{Introduction}
\label{introduction}

The massive growth in model and dataset sizes for deep learning has made distribution a standard approach to training. 
The most popular strategy is \emph{synchronous data-parallel distribution}, which splits the data between parallel workers, 
each of which computes stochastic gradients over their data, and then averages the workers' gradients in a synchronous step.
This procedure has several advantages, but  induces two main overheads: 
the \emph{synchronization cost} of barrier-like synchronization at every step, and the \emph{communication cost} of exchanging the gradients in an all-to-all fashion. 

There has been significant work on mitigating these overheads. 
Specifically, a popular approach for reducing the cost of gradient communication, which is the main focus of our paper, is \emph{lossy gradient compression}~\cite{seide2014, alistarh2017qsgd,dryden2016communication,vogels2019powersgd}, which reduces the number of communicated bits per iteration. 
{Hundreds} of such techniques have been proposed, which can be roughly categorized into {three method families.}
The first general approach is \emph{quantization}~\cite{seide2014, alistarh2017qsgd,wen2017terngrad}, which reduces the bit width of the communicated gradients in a variance-aware fashion, in order to preserve convergence. The second is \emph{sparsification}~\cite{strom2015scalable,dryden2016communication,lin2017deep}, reducing the number of gradient components updated at every step, which are chosen via various saliency metrics. 
The third and most recent approach is \emph{low-rank approximation}~\cite{wang2018atomo,vogels2019powersgd}, which leverages the low-rank structure of gradient tensors to minimize communication cost.

In practice, these approaches come with non-trivial trade-offs in terms of compression versus ease-of-use. 
For instance, gradient quantization is easy to implement and deploy, but only provides limited compression before accuracy degradation; 
sparsification and low-rank approximation can provide order-of-magnitude compression improvements, but come with additional costs in terms of maintaining error correction and careful hyper-parameter tuning for the best results. 
The above trade-offs have been thoroughly investigated by the community, along with new \emph{adaptive} compression methods~\cite{agarwal2021adaptive, CGX2022, faghri2020adaptive}, which use ``learned'' compression, adjusting the compression to the error incurred during various phases of DNN training. 

Despite this vast amount of research, the interplay between these methods, their systems implementation, and the underlying training dynamics, has received significantly less attention. 
Specifically, almost all existing works view the DNN model as a \emph{uniform set of parameters}, and apply compression either \emph{globally} to the entire model, e.g. by performing Top-K selection over the gradients when sparsifying~\cite{Chen2018}, or \emph{uniformly}, applying the same degree of compression to every layer, independently of layer size or impact on the loss, i.e. ``sensitivity''~\cite{vogels2019powersgd}.   
This view can lead to two main fallacies. 
First, from the application side, this misses significant opportunities for optimization: 
 modern deep models, in particular  Transformers~\cite{vaswani2017attention} can be highly heterogeneous in terms of both layer sizes 
 and layer sensitivity to gradient compression, and gradient compression can have a different impact during stages of training~\cite{achille2018critical}. 
Second, from the systems side, most efficient training frameworks overlap communication and computation, transmitting layer gradients as soon as they are generated.  Thus, obtaining a consistent global view of parameters, 
necessary to perform global Top-K selection is either impossible, or very costly to implement and apply.  

These concrete examples point to a significant gap between algorithmic compression techniques, 
necessary for efficient data-parallel training, and their practical implementations. 
More concretely, it is natural to ask: 
given an arbitrary model and a compression technique, is there an efficient way to balance the application constraints, i.e. the \emph{layer sensitivities}, on the one hand, 
and the system communication constraings, i.e. \emph{layer sizes}, on the other, dynamically during training, in order to maximize speedup, without sacrificing final model accuracy?

To address these questions, we introduce \texttt{\projectName}, an efficient and general framework for Layer-wise parametrization of  GRadiEnt COmpression schemes. 
Algorithmically, L-GreCo is based on a new formalization of the \emph{layer-wise adaptive compression} problem, which identifies per-layer compression parameters, e.g. per-layer sparsity settings or quantization levels, maximizing the amount of compression, under a fixed constraint on the total error due to gradient compression, set so there is not accuracy loss. 
At the system level, \texttt{\projectName} works by integrating with standard training frameworks, such as \texttt{torch.distributed}, to exploit model heterogeneity in terms of both per-layer structure and per-layer sensitivity, determining on-the-fly by how much to compress individual layer gradients in order to maximize compression or end-to-end training times. 

We validate \texttt{\projectName} across \emph{all existing families} of compression strategies: quantization, sparsification, and low-rank compression, and across a variety of vision and language tasks. 
Our experiments show that \texttt{\projectName} consistently achieves higher compression rates than existing manual or adaptive strategies, and is particularly-effective for heterogeneous and sensitive Transformer-based models. 
Specifically, the framework provides gains for all existing compression strategies in a black-box fashion, and results in significant compression and practical speedup gains, compared to the best-known manual parametrization schemes, when executing standard image classification or language modeling benchmarks in single- and multi-node settings.


We summarize our contributions as follows: 
\begin{itemize}[topsep=0pt,itemsep=-1ex,partopsep=1ex,parsep=1ex]
    \item We show that gradient compression schemes can take advantage of the heterogeneous layer-wise structure of DNNs in order to reduce communication overhead.
    \item We design and implement the \texttt{\projectName} framework, which guarantees optimal layer-wise compression levels in terms of the compression-accuracy tradeoff, based on a theoretically-justified measure of layer compression sensitivity. 
    \item We provide an extensive empirical evaluation on different neural networks (ResNet18, ResNet50, Transformer-XL, Transformer-LM) with different datasets (CIFAR-100, ImageNet, Wikitext-103) showing that \texttt{\projectName} reduces communication by up to 5$\times$ and achieves speedups up to 2.5$\times$ without loss of accuracy or significant tuning, on single and multi-node deployments.
    \item We conduct the first detailed study for both layer sensitivity metrics and for performance metrics. 
    We find that sensitivity metrics based on quantization error are quivalent to metrics based on output loss. 
    At the same time, from the performance perspective, metrics which seek to maximize the compression ratio lead to similar results to timing-based ones.
    \item Finally, we show that \texttt{\projectName} 
    is compatible with other adaptive compression schemes: specifically, it can be extended to use information about the different stages of training~\cite{agarwal2021adaptive}, leading to further performance improvements. 
\end{itemize}

\begin{figure}
     \centering
     \includegraphics[width=0.5\textwidth]{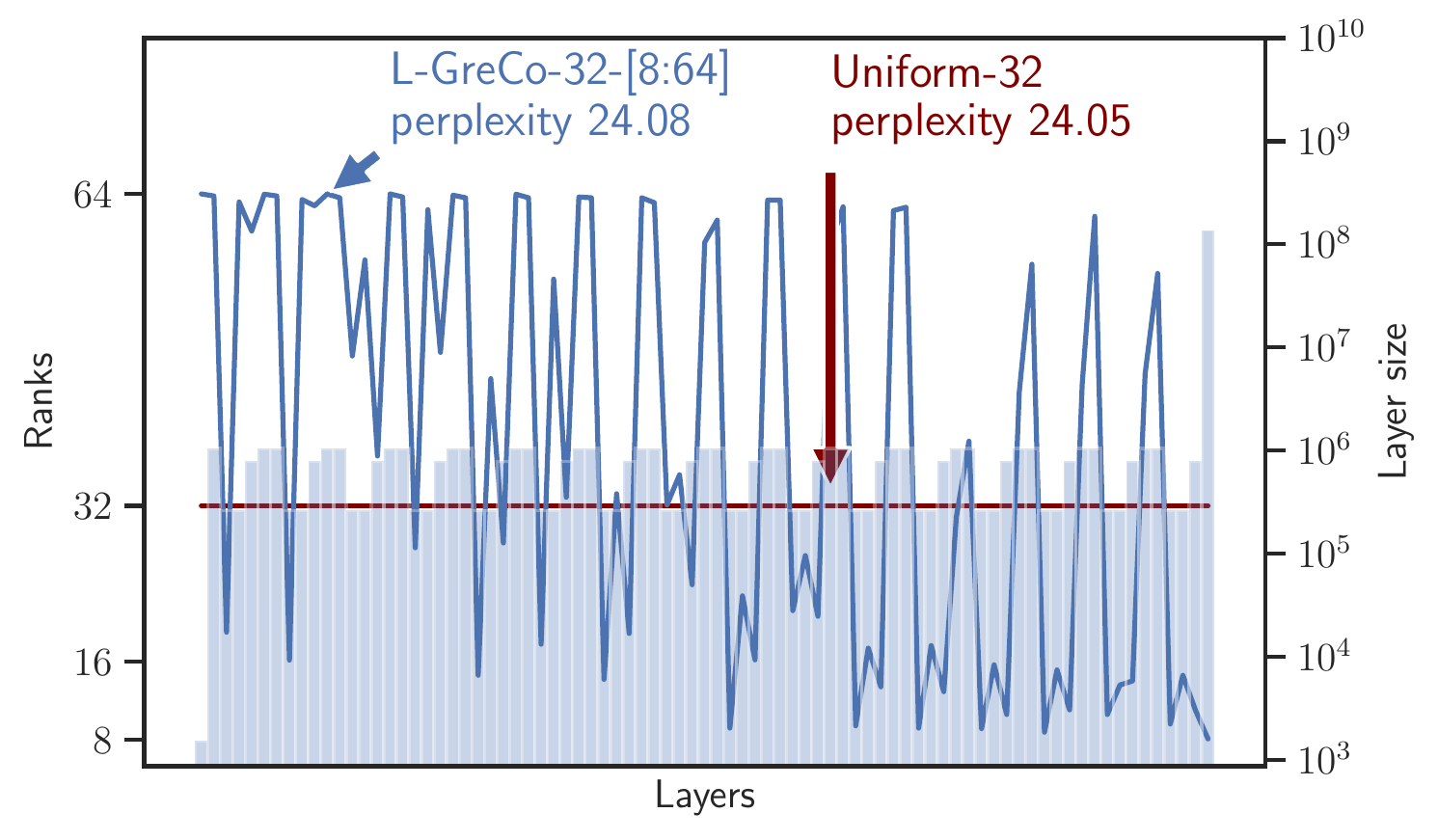}
     \caption{Profile of \texttt{\projectName} rank choices for PowerSGD compression on Transformer-XL. The red line represents uniform compression, while the blue line represents the \texttt{\projectName} profile. Transparent bars show layer sizes. Layers are indexed in the order they are communicated. The annotated number is the final test perplexity (ppl) for the experiment (lower is better). Here, the average compression of \texttt{\projectName} is \textbf{1.5x higher} than uniform.}
     \label{fig:txl_layers}
 \end{figure}

\section{Related work}

\paragraph{Compression methods.}
Gradient compression usually employs three strategies: 
quantization, sparsification, and low-rank decomposition. 
Quantization methods~\cite{seide2014, alistarh2017qsgd, wen2017terngrad, lim20183lc, ramezani2021nuqsgd} use lower precision of each gradient component,  reducing the number of transmitted bits. 
They are easy to implement, and work under stable hyper-parameters~\cite{alistarh2017qsgd, xu2021grace, CGX2022}; however, their compression is limited by the fact that at least one bit per entry must usually be transmitted. 
Sparsification techniques~\cite{strom2015scalable, dryden2016communication, lin2017deep, karimireddy2019error} circumvent this by identifying \emph{salient components} in the gradient and only transmit such subsets. 
Finally, gradient decomposition algorithms~\cite{vogels2019powersgd, wang2018atomo} use the fact that the layer-wise gradient tensors are known to be well-approximable via low-rank matrices, and aim to design light-weight projection approaches that also provide low error. 
Sparsification and low-rank techniques usually require \emph{error correction} buffers to preserve good convergence, as well as non-trivial hyper-parameter tuning. 
As we show experimentally, \texttt{\projectName} is compatible with all of these approaches and can provide significant additional bandwidth savings for each such strategy, without sacrificing model accuracy and without tuning.

\paragraph{Adaptive Compression.}
The general idea of \emph{adapting} the degree of compression during training has been investigated by AdaComp~\cite{Chen2018}, which proposes a self-tuning adaptive compression method; yet, their method does not adapt compression parameters per layer, and cannot be combined with other compression approaches. 
\citet{faghri2020adaptive} adapts the quantization grid to the gradient distribution; yet, their approach is specifically-tuned to quantization, and oblivious to layer heterogeneity.

 \citet{sahu2021rethinkinggs} optimize \emph{the total error over steps} for \break sparsification-based compression and suggest threshold global sparsifiers, which are shown to reach higher compression rates than uniform per-layer compression on small vision tasks (e.g. ResNet18 on CIFAR10/100). However, their approach is restricted to sparsification and leaves unclear how to tune the threshold for large-scale, sensitive models such as Transformers or ImageNet-scale models. In particular, we were unable to obtain good results with this approach on models such as Transformer-XL or Transformer-LM. 
 In section~\ref{sec:accordion} we present a comparison with their approach on ResNet18/CIFAR-100, showing that our method yields both higher accuracy and higher compression.

Accordion~\cite{Agarwal2021} adapts the compression parameters of sparsification and low-rank compression based on the critical regimes of training. 
The algorithm alternates between two compression levels (``low'' and ``high''),  provided by the user and is prone to accuracy loss.
Our approach improves upon Accordion in terms of speedup, but also that we can combine our method with Accordion and obtain even higher gains. 
A variant of the \emph{layer-wise} compression problem similar to the one we consider here was proposed by CGX~\cite{CGX2022}. 
However, they investigate a kmeans-based heuristic, which we show experimentally to be sub-optimal. 

To our knowledge, our dynamic programming strategy has not been employed in the context of adaptive gradient compression. 
Related approaches have been investigated in the context of weight compression for DNNs, see e.g.~\cite{wu2020constraint, aflalo2020knapsack, frantar-spdy, shen2022structural}. Yet, there are significant technical differences: first, the error metrics and speedup objectives are different in the case of weight compression; second, we execute online, at training time, which means that our algorithm has to be extremely efficient, and adapt to dynamic inputs.


\section{Problem Setup}
\label{sec:problem_setup}

\paragraph{Goals.} 
Assuming we are given a DNN model $\mathcal{M}$ with $L$ layers and a compression technique, 
we would intuitively like to find a choice of compression parameters $c_{\ell}$, one for each layer $\ell \in \{1, 2, \ldots, L\}$ which would minimize a metric representing a damage of the training quality introduced by compression while minimizing the total number of bits transmitted.
Yet, this intuitive description leaves open a range of details, such as 
1) the notion of layer-wise metric that corresponds to the compression effect for a given set of parameters; 
2) the exact problem formulation, constraining the compression affect or the compression ratio; and 
3) an efficient implementation of such an algorithm. We address these questions next. 

\paragraph{Metric.} 
\label{sec:metric}
Choosing the right sensitivity metric is a key for accuracy recovery. 
For this, we have tried several approaches. 

First, we note that the sensitivity of a layer to the gradient compression can be measured by a model loss reaction to the compression. 
In order to evaluate the metric we setup the following experiment. We saved model checkpoints at different stages of the uncompressed training. Then conducted multiple short (up to 50 steps) training runs with the same data starting from the checkpoint but varying gradient compression parameters. After that, we used the difference of loss with and without compression as metric. 

Now, the question is what compression parameters (vectors of the size equal to number of the compressed layers) we want to use to evaluate each layer's metric. As long as we wanted to see the model reaction to individual layer compression, for each layer we varied compression parameter leaving all other layers' gradients untouched. With that we collect the differences of loss for each layer and compression parameter and use them as a metric.
Another approach is based on prior theoretical works, which suggest that the \emph{squared $\ell_2$ error of compression} is a good measure of the convergence impact of compression technique. Here, we aggregate gradients during the training, then compress the aggregated gradients for each compression parameter for each layer individually and use the magnitude of the error as a metric.

As shown in Section~\ref{sec:experiments_metric}, the two approaches are strongly correlated and the resulting optimal parameters are close to each other. However, the loss-based approach is not applicable in the real world as it requires \textit{off-training} evaluation which alters the original training pipeline and takes a time comparable to the original training time. At the same time, the error-based approach can be used \textit{during} the training easily integrates into the training and has negligible timing overheads.
Taking this into account, throughout the paper we use L2 norm of the compression error as the main layer sensitivity metric. 

\paragraph{The Constrained Optimization Problem.} 
We formalize our optimization problem as follows. 
Given a model $\mathcal{M}$ with $L$ layers $\ell \in \{1, 2, \ldots, L\}$ and a compression technique, 
providing a set of compression choices $\mathcal{C} = \{c^1, c^2, \ldots c^k\}$ for each layer. 
We emphasize that, for simplicity, we consider a single compression technique and the same compression choices/levels for each layer, but our approach would also work for different techniques being applied to the same model, and heterogeneous compression choices.  

In this context, our method receives as input an error function $\id{error}(\ell, c^j)$ which provides the L2 norm of the compression error at layer $\ell$ for compression choice $c^j$, and a function $\id{size}(\ell, c^\ell)$ which measures the transmission cost of layer $\ell$ for choice $c^\ell$. 
In addition, we assume to be given a fixed maximal error threshold $\mathcal{E}_{\max}$ which the algorithm should not violate. 
Then, we wish to find a layer-wise setting of compression parameters $c_1, \ldots, c_L$ with the objective 

\begin{equation*}
    \label{eqn:objective} 
    \textnormal{ minimize } \sum_{\ell = 1}^L \id{size}(\ell, c^\ell) \, \textnormal{ s.t. } \sum_{\ell = 1}^L \id{error}(\ell, c^\ell) \leq  \mathcal{E}_{\max}. 
\end{equation*}

In practical terms, this formulation aims to minimize the total transmission cost for the gradient tensors under a maximum \emph{additive} constraint on the gradient compression error. One implicit assumption is that the metric $\id{error}(\cdot, \cdot)$ is \emph{additive} over layers, and that it is possible to obtain a ``reference'' error upper bound which does not result in accuracy loss. As we see next, this is the case for the error metric we adopt.

\paragraph{The Error Bound.}
The remaining question is how to choose $\mathcal{E}_{\max}$. 
We decided to pick this error bound to track that of a reference compression approach which is \emph{known not to lose accuracy relative to the baseline}. 
Here, we leverage the fact that the literature provides parameters which allow reaching full accuracy recovery for different models and datasets. 
For instance, for quantization, we can use \emph{4-bit quantization}, which is known to recover for virtually every model~\cite{alistarh2017qsgd, CGX2022}. For sparsification ~\cite{lin2017deep, renggli2019sparcml} as well as for matrix decomposition~\cite{vogels2019powersgd} we had to use different reference parameters for different training according to their baselines (For the details refer to the Tables ~\ref{table:results_ic} and~\ref{table:results_lm}).

\begin{figure}[t]
\centering
\includegraphics[width=0.4\textwidth]{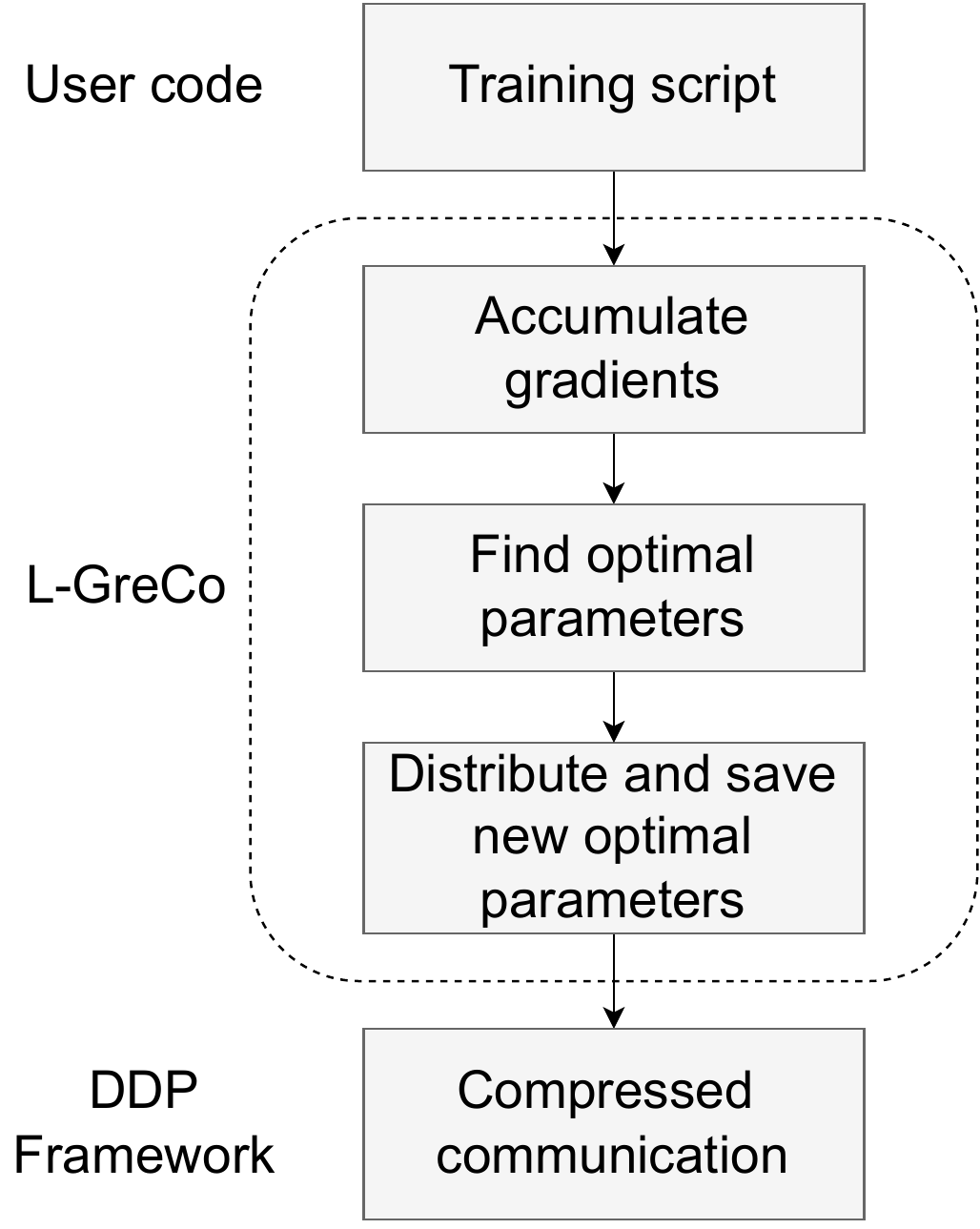}
\caption{Schema of \projectName~integration.}
\label{fig:diagram}
\end{figure}

\section{The \projectName\ Framework}

\paragraph{Overview.} 
We now describe a general algorithm to solve the constrained optimization problem from the previous section. Our algorithm makes layer-wise decisions in order to balance the magnitude of the compression error and the compressed size of the model. 
As inputs, our algorithm takes in the uncompressed layer sizes $\id{size}(\ell, \bot)$, a set $\mathcal{G}$ of accumulated gradients per layer (which will be used to examine compression error), as well as a fixed error bound $\mathcal{E}_{\max}$. 
Specifically, at a given decision step, the objective is to find an optimal mapping of each layer $\ell$ to a compression level $c_\ell$, such that the norm of the total compression error, computed over the set of accumulated gradients $\mathcal{G}$ does not surpass $\mathcal{E}_{\max}$, but the total compressed size of the model $\sum_{\ell = 1}^L \id{size}(\ell, c_{\ell})$ is minimal for this error bound. 

This formulation is reminiscent of the \emph{knapsack problem}: the error is the size of the knapsack, and the compressed size is the value we wish to optimize. 
In this formulation, the problem would have an efficient and optimal algorithm, using dynamic programming (DP). 
However, the squared $L_2$ error is not discrete, so we cannot directly apply this approach. 
Instead, the idea is to reduce this to a solvable problem by \emph{discretizing} the possible set of error values. 
Specifically, since errors are monotonic and we can use a very fine discretization without significant efficiency loss, 
it is highly unlikely that we would miss the optimal solution by a significant amount. 
For illustration, in our implementation, we use $D = 10000$ as a discretization factor (i.e. steps of size $\mathcal{E}_{\max} / D$).


\begin{algorithm}[t]
 \caption{\projectName\ adaptive compression}
 \label{algo:dp}
 {\small
 \begin{algorithmic}[1]
 \renewcommand{\algorithmicrequire}{\textbf{Input:}}
 \renewcommand{\algorithmicensure}{\textbf{Output:}}
 \REQUIRE Model Layers $L_i$, accumulated gradients $G_i$, possible compression parameters $C = \{c^1, c^2, \ldots, c^k\}$, static default compression parameters we try to improve $C_i^d$, discretization factor $D$ 
 \ENSURE Compression parameters assignments $c_\ell \in C$ for each layer $\ell$\\
 \STATE N = number of layers
 \STATE Compute $\mathcal{E}_{\max}$ for the default compression parameters $C_i^d$
 \STATE Compute discretization step $\mathcal{E}_{\max} / D$.
 \STATE Costs matrix $N \times |C|$ where position $i, j$ has a value of the size of layer $i$ compressed with compression parameter $c^j$.
 \STATE Errors matrix $N \times |C|$ where position $i, j$ has a value of the discretized $L_2$ of the compression error when the accumulated gradients of layer $i$ are compressed with parameter $c^j$.
 \STATE $DP$ matrix $N \times (D + 1)$ filled with $\infty$ values.
 \STATE $PD$ matrix $N \times (D + 1)$.
 \STATE // Initialization of the cost tables: 
 \FOR{$c \in C$} 
    \STATE $DP[1][Errors[1][c]] = Costs[1][c]$
    \STATE $PD[1][Errors[1][c]] = c$
 \ENDFOR
 \STATE // Dynamic programming algorithm
 \FOR{Layer $l_i:= 2 .. N$} 
    \FOR {$c_i \in C$}
        \FOR {$e_i:= Errors[l_i][c_i] .. D$}
            \STATE $t = DP[l_i - 1][e_i - Errors[l_i][c_i]] + Costs[l_i][c_i]$
            \IF{$t < DP[l_i][e_i]$}
                 \STATE $DP[l_i][e_i] = t$
                \STATE $PD[l_i][e_i] = c_i$
            \ENDIF
        \ENDFOR
    \ENDFOR
 \ENDFOR
 \STATE $err_{min} = argmin(DP[N])$
 \STATE // Reconstruction of the optimal parameters
 \FOR{$l_i = N .. 1$}
    \STATE $result[l_i] = PD[l_i][err_{min}]$
    \STATE $err_{min} = err_{min} - Errors[l_i][result[l_i]]$
 \ENDFOR
 \STATE \textbf{return} $result$
 \end{algorithmic}
 }
\end{algorithm}

\paragraph{The Algorithm.} 
\label{sec:algorithm}
The algorithm, presented in full in Algorithm~\ref{algo:dp}, works as follows.
First, we compute the data needed for the algorithm for all layers and all considered compression parameters (lines 1-10), corresponding to errors and compressions for each possible choice. 
Then we execute a dynamic programming algorithm solving the following problem.
We want to compute the minimum total size given total compression error $E$ in the first $\ell$ layers $\id{compressed size}(\ell, E) = \min_{\ell} \id{compressed size}(\ell - 1, E - error(\ell, c_{\ell})) + size(\ell, c_{\ell})$. To achieve this, for each layer we want to consider, we run over all error increments and all possible compression parameters, and minimize the total compressed size for the current total compression error (lines 12-22) saving the compression parameter with which we obtain the minimum. Then, in lines 23-27, we find the error increment with which we achieve the lowest total compressed size and reconstruct the compression parameter mapping---thus, we have obtained the result.

\paragraph{Implementation.} 
As shown in the Figure.~\ref{fig:diagram} we integrate the \texttt{\projectName} in the middle between user training code and communication system that is responsible for compression of the gradient and their synchronization.
We run the above algorithm periodically, e.g. once per training epoch, on a single designated worker; unless otherwise stated, this worker performs all steps. 
In between runs of the algorithm, we accumulate per-layer gradients in auxiliary buffers. We then build an $L_2$ error table, for each layer, and for every compression parameter in the user-provided range and for the reference compression parameters set. 
To find the error, we simulate the compression/decompression of each layer with the given compression parameter without applying error feedback and compute the $L_2$ distance between the original and recovered vectors. Then we run the DP algorithm. This provides us with the optimal compression mapping, which the designated worker broadcasts to the other workers. Then, on each worker, we save the compression parameters mapping in the communication engine.

\paragraph{Computational and memory costs.}
The algorithm assumes that we accumulate gradients in additional buffers, occupying the model size memory.
The DP algorithm has $O(D|L||C|)$ time complexity and $O(|L|D)$ memory complexity. The actual timings for the algorithm are presented in Table~\ref{table:overheads}. The overheads consist of two parts: 1. Error computation 2. Dynamic programming algorithm. We can see that the dynamic programming part only takes a fraction of a percent of the training time whereas most of the overhead is caused by computation. However, both overheads are negligible compared to the speedups provided by \texttt{\projectName} (see Figures ~\ref{fig:txl_multi} and ~\ref{fig:rn50_multi}). 


\begin{table}[h]
\centering
\caption{ Timing overheads for \texttt{\projectName} in relation to the total training time. Numbers in brackets represent error computation.}
\label{table:overheads}
{\footnotesize
\begin{tabular}{|M{15mm}|M{25mm}|M{25mm}|}
\hline
Compression Method & Transformer-XL & ResNet50\\
\hline
PowerSGD & $0.56\% [0.49\%]$ & $0.15\% [0.14\%]$ \\
\hline
QSGD & $0.14\% [0.13\%]$ & $0.04\% [0.03\%]$ \\
\hline
TopK & $0.38\% [0.35\%]$ & $0.33\% [0.30\%]$ \\
\hline

\end{tabular}
}
\end{table}
\paragraph{Communication details.}
\label{sec:comm_details}
From the point of view of the common data-parallel implementation, the gradients become available right after the end of the backward propagation of the corresponding layer. In order to reduce communication latency, gradients are grouped into several buffers (called buckets in \texttt{pytorch}). An important optimization is the overlapping of gradient communication with computation. This implies that the cost of the communication of the first buckets is likely to be completely ``hidden'' by computation, whereas the synchronization of the last bucket becomes a significant part of the timing delay between the training steps (see Figure.~\ref{fig:global_topk}(b)). This means that the communication speed of different parts of the model (i.e., layer groups) has a different impact on the training speed. Hence, compression ratio optimization might output suboptimal results.

\paragraph{Time optimization.}
\label{sec:time_optimization}
Having that in mind, we have integrated an utility that measures time each bucket takes to synchronize. With this tool, we can reformulate an optimization problem: in fact, we want to minimize the time range between the start of first and end of the last bucket synchronization (i.e. gradient synchronization time) rather than the compression ratio. For that, we measure gradients synchronization time for various combinations of compression ratios per bucket, saving the communicated buckets sizes.
Then, we train a linear regression model  to learn the relation between the transmitted bucket sizes and the gradient synchronization time. From that we obtain per-bucket coefficients $T(b)$ that we can apply for each layer in a respective bucket.
Then we change the objective in the optimization problem(see Formula.~\ref{eqn:objective}) to: 
\begin{equation*}
    \label{eqn:objective_time} 
    \textnormal{ minimize } \sum_{\ell = 1}^L \id{size}(\ell, c^\ell) * T(\ell) \, \textnormal{ s.t. } \sum_{\ell = 1}^L \id{error}(\ell, c^\ell) \leq  \mathcal{E}_{\max}.
\end{equation*}

This gives us the parameters that have the same or worse compression characteristics than ratio-based algorithm but improving the estimated communication time.

\section{Experimental Validation}
\label{sec:experiments}
We experimentally evaluate \texttt{\projectName} across all existing compression strategies: quantization using QSGD, TopK sparsification, and low-rank approximation via PowerSGD. 


\subsection{Experimental setup}
\label{sec:exp_setup}
\paragraph{Infrastructure.} Our evaluation uses commodity workstations with 4 or 8 NVIDIA RTX3090 GPUs.
In the multi-node setting, we use 4 cloud instances with 4xRTX3090 GPUs, provided by Genesis Cloud. Bandwidth measurements show that inter-GPU bandwidth values lie in between 13 to 16 Gbps, and inter-node bandwidth in the cloud is up to 10 Gbps. We used Pytorch 1.10, openmpi/4.1.4, CUDA 11.3, NCCL 2.8.4, and cudnn/8.1.1.

\paragraph{Implementation.}
We implement \texttt{\projectName} in PyTorch using \texttt{torch.distributed} hooks for PowerSGD and leveraging open-source CGX framework~\cite{CGX2022} for quantization and sparsification.


\paragraph{Datasets and models.}
We examine two different DNN learning tasks: 1) image classification on the CIFAR100~\cite{krizhevsky2009learning} and ImageNet~\cite{deng2009imagenet} datasets, and 2) language modeling on WikiText-103~\cite{merity2016pointer}.
We used state-of-the-art model implementations and parameters provided by the PyTorch version of the NVIDIA Training Examples benchmark~\cite{DeepLearningExamples} and the \texttt{fairseq} library PyTorch examples~\cite{ott2019fairseq}. 
We used ResNet-18 for CIFAR-100 training with batch size 256, ResNet-50 in the mixed-precision regime for ImageNet with batch size 2048, and Transformer-XL and Transformer-LM trained in full-precision for WikiText-103, with batch sizes 256 and 2048, respectively. 
\textbf{All our experiments use the original training recipes, without any hyperparameter tuning to account for gradient-compressed training.} 



\paragraph{Baselines.}
The first natural baseline is uncompressed training, which sets our accuracy baseline. Matching MLPerf~\cite{mattson2020mlperf}, we set our accuracy  threshold to 1\% relative to uncompressed training.
The second natural baseline is the \emph{best existing manually-generated} gradient compression recipes. 
By and large, existing methods propose \emph{uniform} per-layer compression to a given threshold, e.g.~\cite{alistarh2017qsgd, wen2017terngrad, renggli2019sparcml, vogels2019powersgd}. 
For such baselines, we want to improve compressed size and training speed, possibly also improving final model accuracy.
We found that the best choice of compression parameters for uniform per-layer assignment depends on the compression method, dataset, and task. For some experiments we had to tune the uniform compression parameters to match baseline (non-compressed) results (see Tables~\ref{table:results_ic} and ~\ref{table:results_lm}). 
 

 \begin{table*}[t]
 \centering
 
 \caption{Accuracy recovery and compression ratios for different compression methods with uniform and adaptive schemes on image classification tasks. The compression ratios measure actual transmission savings. Values in brackets for \texttt{\projectName} compression ratios stand for improvements relative to the corresponding uniform compression.}
 \label{table:results_ic}
\begin{tabular}{|P{18mm}|P{20mm}|lll|lll|}
\hline
Compression approach & Parameter choice & \multicolumn{3}{l|}{ResNet18 on CIFAR-100} & \multicolumn{3}{l|}{ResNet50 on ImageNet} \\ \hline
 &  & \multicolumn{1}{p{10mm}|}{Default param} & \multicolumn{1}{P{17mm}|}{Accuracy} & \multicolumn{1}{P{17mm}|}{Compression ratio} & \multicolumn{1}{p{10mm}|}{Default param} & \multicolumn{1}{P{17mm}|}{Accuracy} & \multicolumn{1}{P{17mm}|}{Compression ratio} \\ \hline
Baseline & N/A & \multicolumn{1}{P{10mm}|}{-} & \multicolumn{1}{l|}{$76.60 \pm 0.40$} & \multicolumn{1}{P{18mm}|}{1.0} & \multicolumn{1}{P{10mm}|}{-} & \multicolumn{1}{l|}{$76.88 \pm 0.16$} & \multicolumn{1}{P{18mm}|}{1.0} \\ \hline
\multirow{2}{*}{QSGD} & uniform & \multicolumn{1}{P{10mm}|}{\multirow{2}{*}{4 bit}} & \multicolumn{1}{l|}{$76.80 \pm 0.40$} & \multicolumn{1}{P{17mm}|}{7.8} & \multicolumn{1}{P{10mm}|}{\multirow{2}{*}{4 bit}} & \multicolumn{1}{l|}{$77.38 \pm 0.10$} &  \multicolumn{1}{P{17mm}|}{7.7} \\ \cline{2-2} \cline{4-5} \cline{7-8} 
 & L-Greco & \multicolumn{1}{l|}{} & \multicolumn{1}{l|}{$76.46 \pm 0.21$} & \multicolumn{1}{P{18mm}|}{8.6~$[1.10\times]$} & \multicolumn{1}{l|}{} & \multicolumn{1}{l|}{$76.77 \pm 0.25$} & \multicolumn{1}{P{17mm}|}{11.0~$[1.41\times]$} \\ \hline
\multirow{2}{*}{TopK} & uniform & \multicolumn{1}{P{10mm}|}{\multirow{2}{*}{1\%}} & \multicolumn{1}{l|}{$75.73 \pm 0.46$} & \multicolumn{1}{P{17mm}|}{48.1} & \multicolumn{1}{P{10mm}|}{\multirow{2}{*}{1\%}} & \multicolumn{1}{l|}{$76.85 \pm 0.06$} &  \multicolumn{1}{P{17mm}|}{45.6} \\ \cline{2-2} \cline{4-5} \cline{7-8} 
 & L-Greco & \multicolumn{1}{l|}{} & \multicolumn{1}{l|}{$75.66 \pm 0.35$} & \multicolumn{1}{P{17mm}|}{182.0~$[3.78\times]$} & \multicolumn{1}{l|}{} & \multicolumn{1}{l|}{$77.04 \pm 0.27$} &  \multicolumn{1}{P{17mm}|}{122.0~$[2.67\times]$} \\ \hline
\multirow{2}{*}{PowerSGD} & uniform & \multicolumn{1}{l|}{\multirow{2}{*}{rank 4}} & \multicolumn{1}{l|}{$76.36 \pm 0.28$} & \multicolumn{1}{P{17mm}|}{72.2} & \multicolumn{1}{l|}{\multirow{2}{*}{rank 4}} & \multicolumn{1}{l|}{$76.50 \pm 0.37$} &  \multicolumn{1}{P{17mm}|}{66.5} \\ \cline{2-2} \cline{4-5} \cline{7-8} 
 & L-Greco & \multicolumn{1}{l|}{} & \multicolumn{1}{l|}{$76.43 \pm 0.37$} & \multicolumn{1}{P{17mm}|}{133.9~$[1.85\times]$} & \multicolumn{1}{l|}{} & \multicolumn{1}{l|}{$76.33 \pm 0.27$} & \multicolumn{1}{P{17mm}|}{96.2~$[1.44\times]$} \\ \hline
\end{tabular}
\end{table*}
 
\begin{table*}[t]
\centering
 
\caption{Accuracy recovery and compression ratios for different compression methods with uniform and adaptive schemes on language modeling tasks. The compression ratios measure actual transmission savings. Values in brackets for \texttt{\projectName} compression ratios stand for improvements relative to the corresponding uniform compression.}
\label{table:results_lm}
\begin{tabular}{|P{18mm}|P{20mm}|lll|lll|}
\hline
Compression approach & Compression parameters & \multicolumn{3}{l|}{TransformerXL on WIKITEXT-103} & \multicolumn{3}{l|}{TransformerLM on WIKITEXT-103} \\ \hline
 &  & \multicolumn{1}{P{10mm}|}{Default param} & \multicolumn{1}{P{17mm}|}{Perplexity} & \multicolumn{1}{P{17mm}|}{Compression ratio} & \multicolumn{1}{P{10mm}|}{Default param} & \multicolumn{1}{P{17mm}|}{Perplexity} & \multicolumn{1}{P{17mm}|}{Compression ratio} \\ \hline
Baseline & N/A & \multicolumn{1}{P{10mm}|}{-} & \multicolumn{1}{l|}{$23.82 \pm 0.10$} & \multicolumn{1}{P{17mm}|}{1.0} & \multicolumn{1}{P{10mm}|}{-} & \multicolumn{1}{l|}{$29.34 \pm 0.12$} & \multicolumn{1}{P{17mm}|}{1.0} \\ \hline
\multirow{2}{*}{QSGD} & uniform & \multicolumn{1}{P{10mm}|}{\multirow{2}{*}{4 bit}} & \multicolumn{1}{l|}{$23.82 \pm 0.1$} & \multicolumn{1}{P{17mm}|}{7.8} & \multicolumn{1}{P{10mm}|}{\multirow{2}{*}{4 bit}} & \multicolumn{1}{l|}{$29.39 \pm 0.10$} & \multicolumn{1}{P{17mm}|}{7.8} \\ \cline{2-2} \cline{4-5} \cline{7-8} 
 & L-Greco & \multicolumn{1}{l|}{} & \multicolumn{1}{l|}{$24.11 \pm 0.09$} & \multicolumn{1}{P{17mm}|}{9.1~$[1.16\times]$} & \multicolumn{1}{l|}{} & \multicolumn{1}{l|}{$30.03 \pm 0.16$} &  \multicolumn{1}{P{17mm}|}{9.9~$[1.26\times]$} \\ \hline
\multirow{2}{*}{TopK} & uniform & \multicolumn{1}{P{10mm}|}{\multirow{2}{*}{10\%}} & \multicolumn{1}{l|}{$24.13 \pm 0.14$} & \multicolumn{1}{P{17mm}|}{4.9} & \multicolumn{1}{P{10mm}|}{\multirow{2}{*}{10\%}} & \multicolumn{1}{l|}{$29.29 \pm 0.09$} & \multicolumn{1}{P{17mm}|}{4.9} \\ \cline{2-2} \cline{4-5} \cline{7-8} 
 & L-Greco & \multicolumn{1}{l|}{} & \multicolumn{1}{l|}{$24.19 \pm 0.13$} & \multicolumn{1}{P{17mm}|}{12.8~$[2.61\times]$} & \multicolumn{1}{l|}{} & \multicolumn{1}{l|}{$29.08 \pm 0.20$} &  \multicolumn{1}{P{17mm}|}{25.6~$[5.2\times]$} \\ \hline
\multirow{2}{*}{PowerSGD} & uniform & \multicolumn{1}{l|}{\multirow{2}{*}{rank 32}} & \multicolumn{1}{l|}{$24.08 \pm 0.12$} & \multicolumn{1}{P{17mm}|}{14.0} & \multicolumn{1}{l|}{\multirow{2}{*}{rank 32}} & \multicolumn{1}{l|}{$29.98 \pm 0.09$} & \multicolumn{1}{P{17mm}|}{15.0} \\ \cline{2-2} \cline{4-5} \cline{7-8} 
 & L-Greco & \multicolumn{1}{l|}{} & \multicolumn{1}{l|}{$24.09 \pm 0.15$} & \multicolumn{1}{P{17mm}|}{20.8~$[1.48\times]$} & \multicolumn{1}{l|}{} & \multicolumn{1}{l|}{$30.19 \pm 0.09$} & \multicolumn{1}{P{17mm}|}{26.5~$[1.76\times]$}  \\ \hline
\end{tabular}
\end{table*}
 

\begin{figure}[t]
\centering
\includegraphics[width=0.49\textwidth]{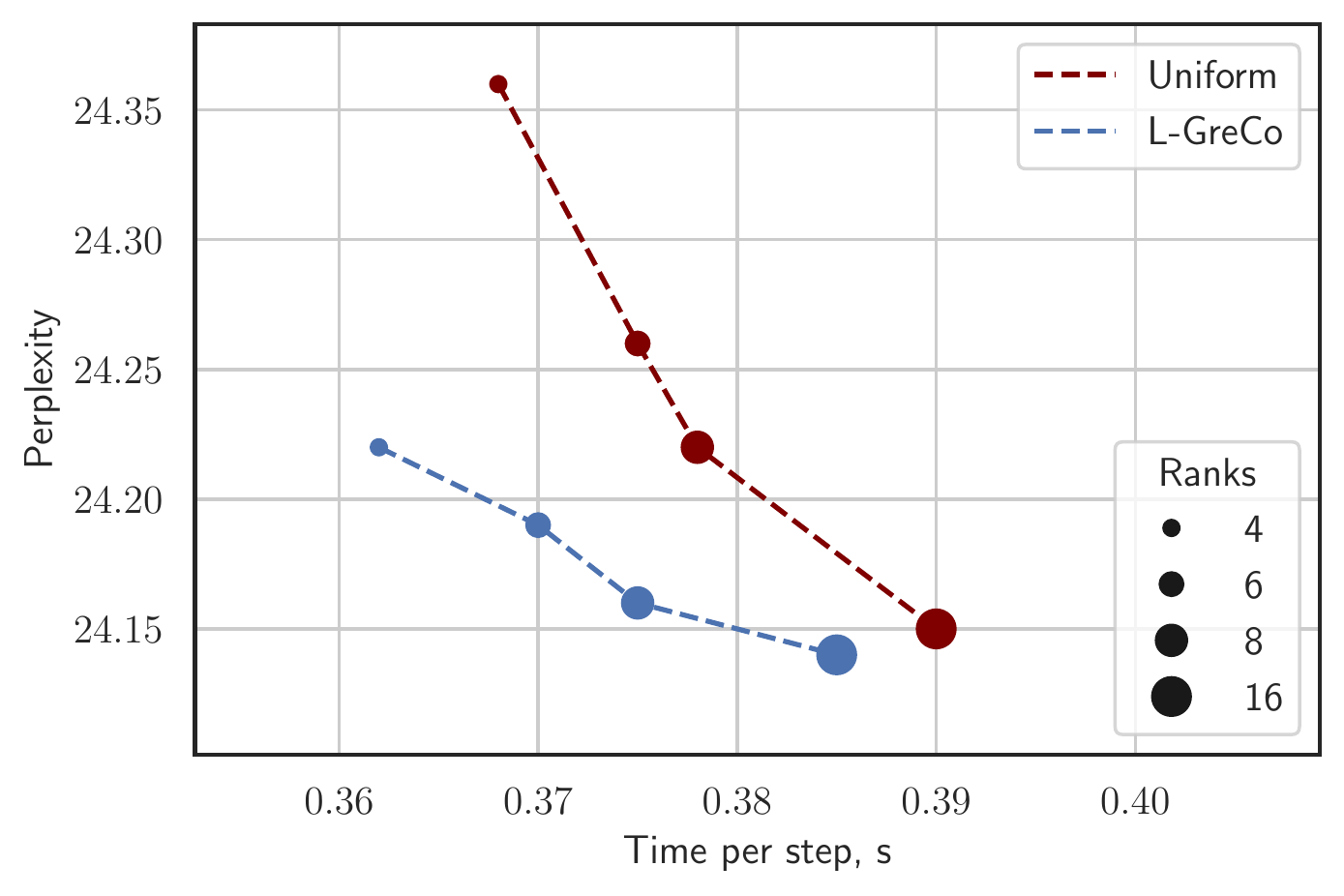}
\caption{Perplexity (lower is better) vs. time per step (smaller is better) for different ranks of PowerSGD compression for the Transformer-XL wikitext-103 task with uniform or \texttt{\projectName} suggested compression schemes. Single node, 8 RTX3090 GPUs.}
\label{fig:txl_psgd_ppl}
\end{figure}

\paragraph{Parameter ranges.}
\texttt{\projectName} requires a range of possible compression parameters as input.
We have always chosen this range to include the default compression parameters used in the literature. Moreover, we left a gap between the default parameter and the  highest possible compression parameter (the right range bound) --- otherwise, we are limited to matching $L_2$ error; and a gap between the default parameter and the lowest possible compression parameter (the left range bound)---otherwise, we will not improve  compression.

The range depends on the compression method. Quantization and low-rank  methods have a limited number of discrete parameters (number of bits per value and decomposition rank respectively) whereas the range for sparsification is larger. In our experiments, we use the following simple approach. 
 Assume a default uniform compression parameter $D$, e.g. 4 quantization bits. For quantization and low-rank methods, the search space was defined as $[D/2, 2 * D]$ with an incremental step of 1. For sparsification, we chose $[D/10, 10*D]$ with an increment of $D/10$.

\begin{figure*}[h]
    \centering
    \setkeys{Gin}{width=0.45\linewidth}
    \subfloat[Compression ratio over time.\label{fig:accordion_ratio_vs_step}]{\includegraphics{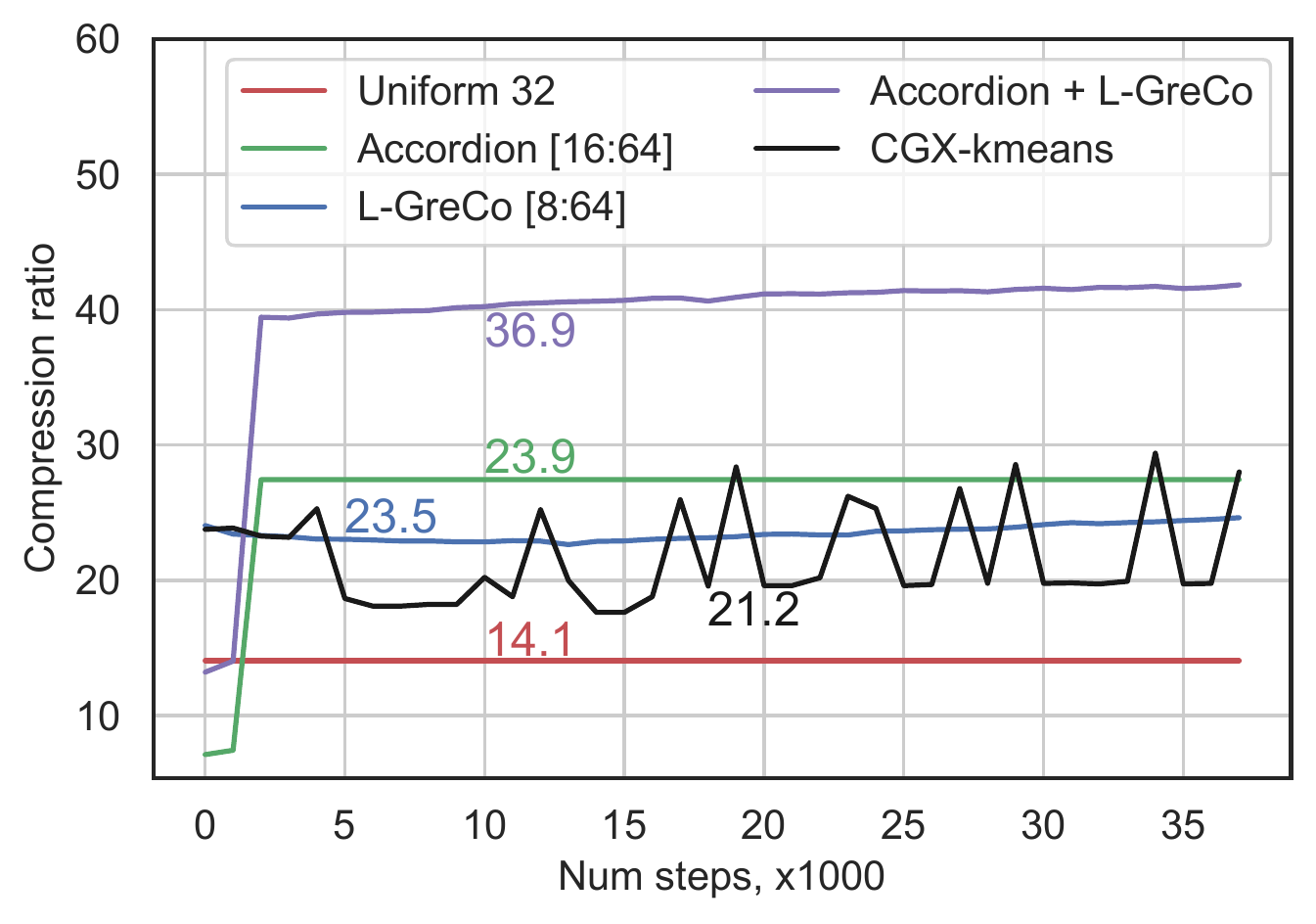}}\hfil
    \subfloat[Compression per layer group (bucket).\label{fig:accordion_buckets}]{\includegraphics{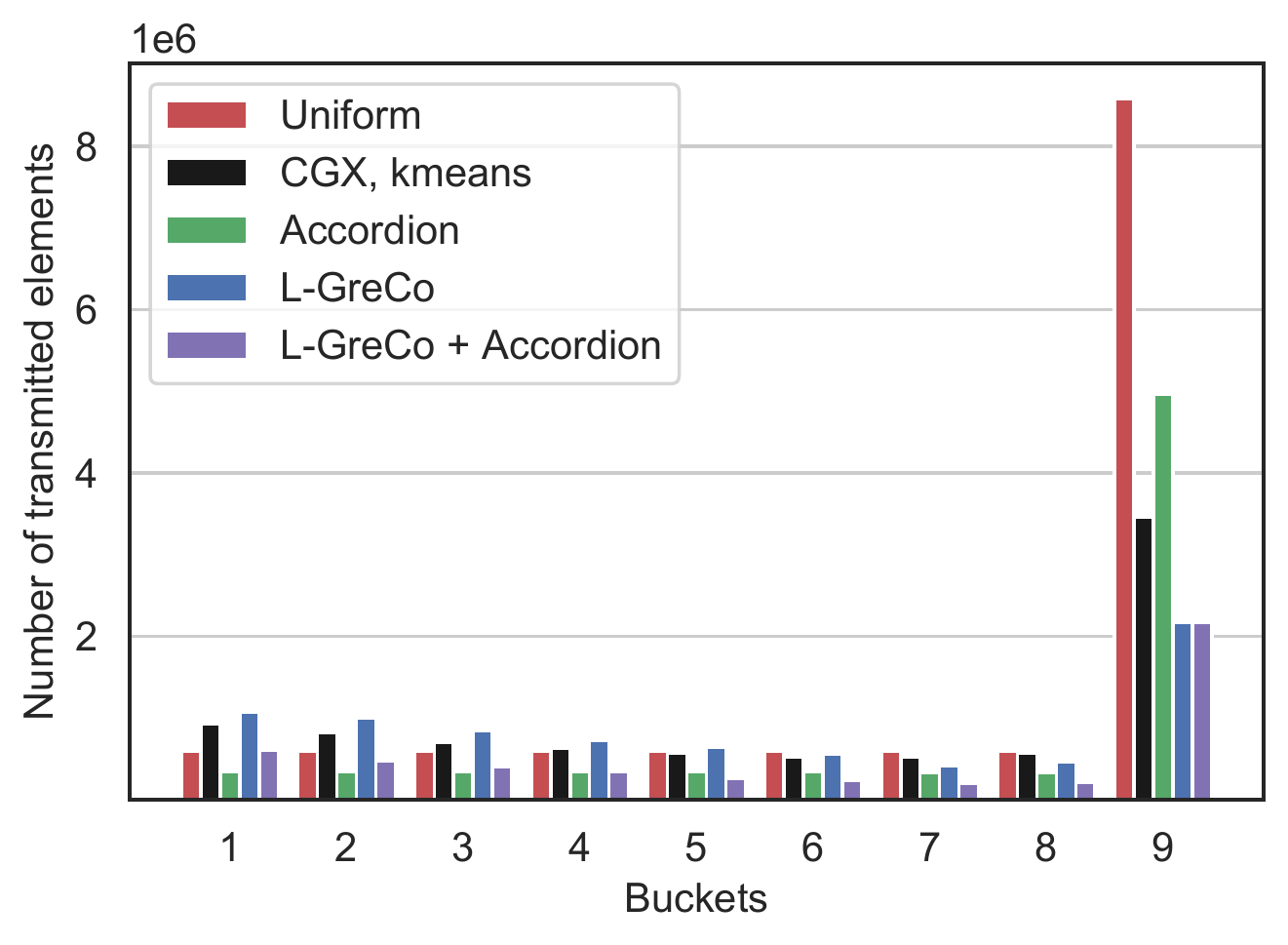}}
    \caption{Adaptive compression using \texttt{\projectName} versus other methods, for PowerSGD compression on Transformer-XL. The left plot shows the dynamics of the compression ratio during training, marking the \emph{average compression ratio}. The right plot presents the transmitted number of elements per bucket averaged over time. Buckets are in communication order.}
    \label{fig:accordion}
\end{figure*}

The other two input parameters of \texttt{\projectName} are how frequently the algorithm is run and the warm-up period after which the compression is turned on. The first parameter is usually chosen to be equal to the evaluation period from the training recipes (typically, 1 epoch). As we will see (Figure~\ref{fig:accordion_ratio_vs_step}) the compression ratio of the schemes returned by \texttt{\projectName} is relatively stable, so it does not need a frequent re-adjustment. In our experiments, the warm-up period is equal to the default learning rate warm-up period.


\subsection{Evaluation results}
\paragraph{Accuracy recovery.}
We first examine model accuracies using standard recipes for  end-to-end training. For each experiment, we performed 3 runs with different seeds.
 We compare \texttt{\projectName} accuracy recovery  with the uncompressed (baseline) and the best uniform per-layer compression parameters (uniform).
 The results are presented in Tables \ref{table:results_ic} and \ref{table:results_lm}.
All accuracies and perplexities are presented with seed variability. The \emph{compression ratio} represents actual transmission cost savings versus the uncompressed baseline. For each compression method and training task, we took the default parameter which provides the highest compression ratio while recovering the final model accuracy, i.e. further compression improvement with uniform setting only leads to \textit{worse} convergence.

 Overall, results show that \texttt{\projectName} stays within the accuracy recovery limit of 1\% multiplicative error~\cite{mattson2020mlperf} for most tasks, often being very close to the uniform baseline, while consistently increasing the compression ratio, across all the tasks and compression techniques considered. 
 We stress that we did not perform task-specific parameter tuning. 
 The gains are remarkably high for sparsification and low-rank techniques, where the search space and therefore also the savings potential of \texttt{\projectName} are higher. For instance, for Transformer-LM, we obtain up to 5x higher compression relative to the uniform baseline, with negligible accuracy impact. At the same time, we note that L-GreCo induces $> 1\%$ multiplicative loss on quantization and low-rank compression for the highly-sensitive Transformer-LM model.\footnote{Specifically, our loss is of at most $0.85$ perplexity relative to the uncompressed baseline. For this model, however, even basic FP16 training loses more than 1 point of perplexity vs FP32.} This is because our default compression range is too aggressive in this case; this can be easily addressed by adjusting the range---we chose not to do it for consistency. 

To further explore the accuracy-compression trade-off, we varied the uniform default compression parameters, specifically throttling the target powerSGD rank for the Transformer-XL/WikiText-103 task.  Figure~\ref{fig:txl_psgd_ppl} shows that \texttt{\projectName} provides a markedly better trade-off than uniform compression. We can see that L-GreCo provides better perplexity recovery while improving training speed of the model.


\begin{figure*}[h]
    \centering
    \setkeys{Gin}{width=0.33\linewidth}    
    \subfloat[PowerSGD\label{fig:txl_multinode_psgd}]{\includegraphics{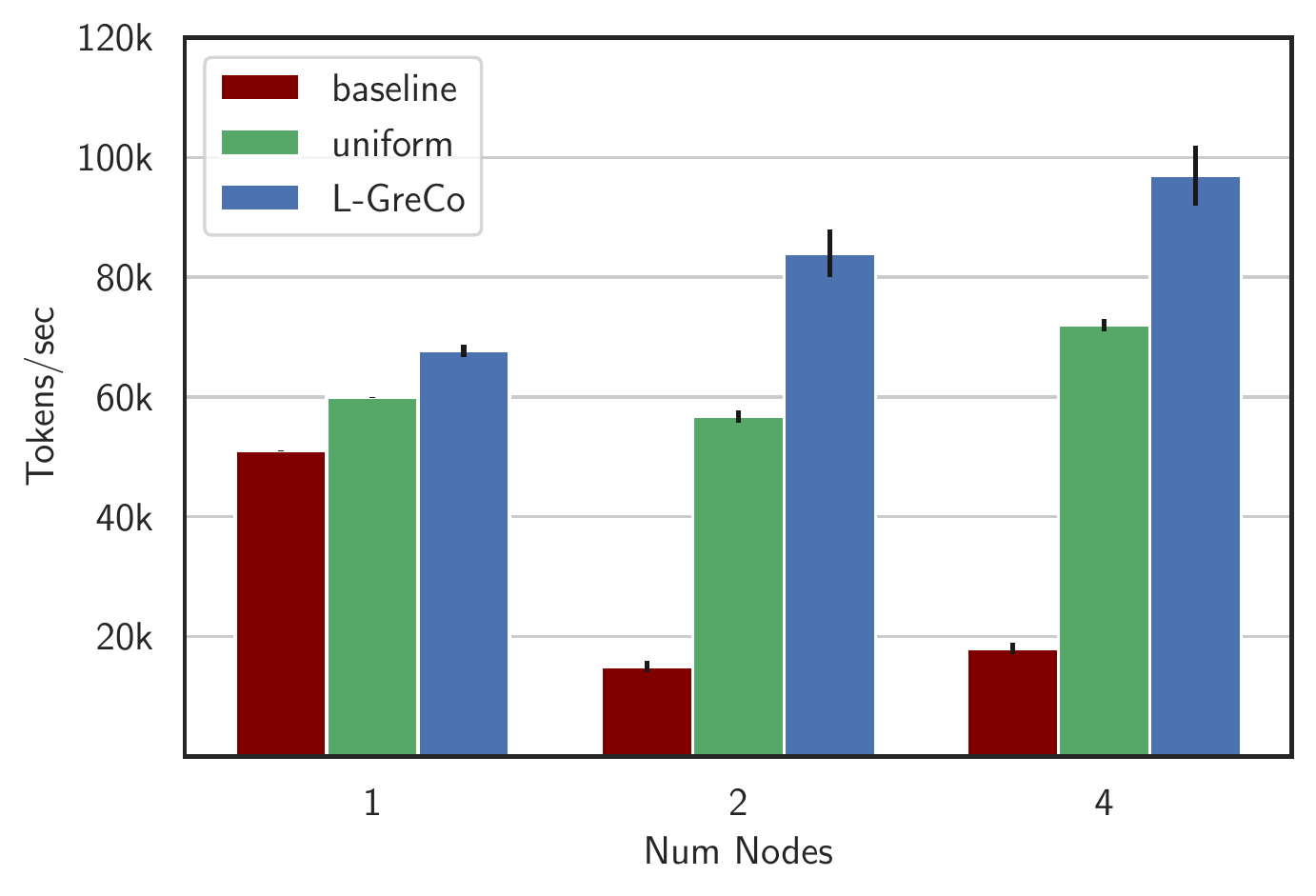}}
    \subfloat[QSGD\label{fig:txl_multinode_qsgd}]{\includegraphics{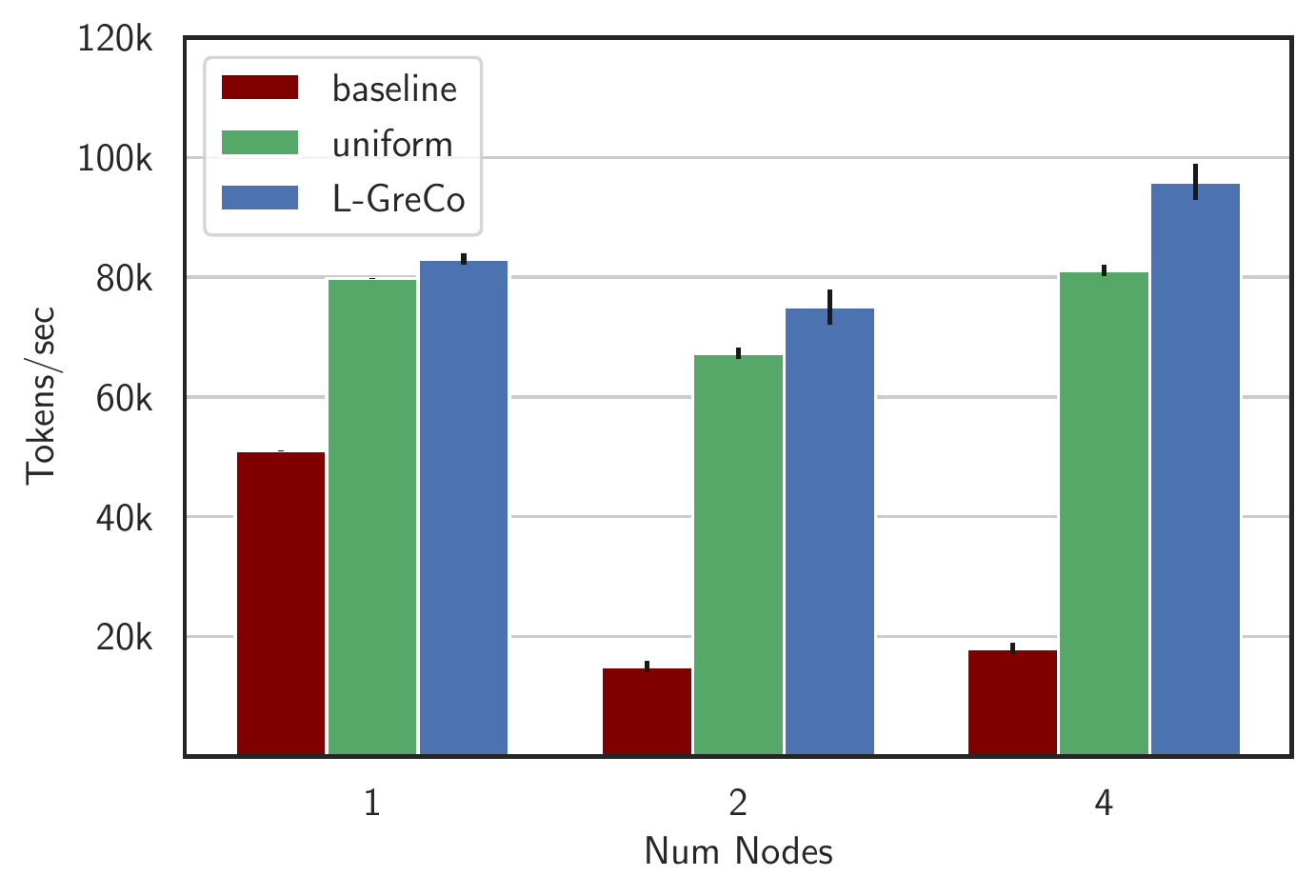}}
    \subfloat[TopK\label{fig:txl_multinode_topk}]{\includegraphics{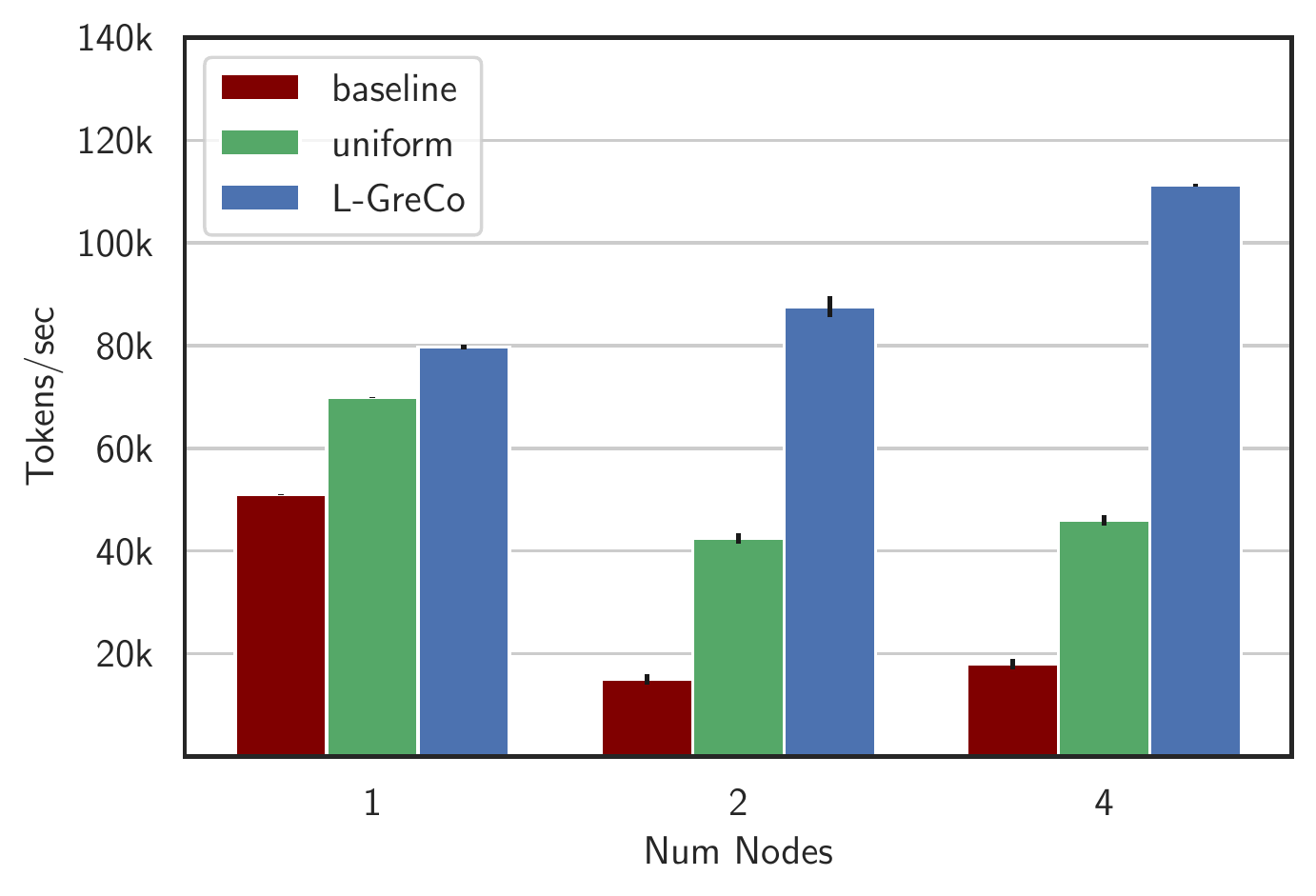}}
    \caption{{\small Throughput for Transformer-XL (TXL) on WikiText-103. Multi node, each node has 4 RTX3090 GPUs.}}
    \label{fig:txl_multi}
 \end{figure*}

\begin{figure*}[h]
    \centering
    \setkeys{Gin}{width=0.33\linewidth}    
    \subfloat[PowerSGD\label{fig:rn50_multinode_psgd}]{\includegraphics{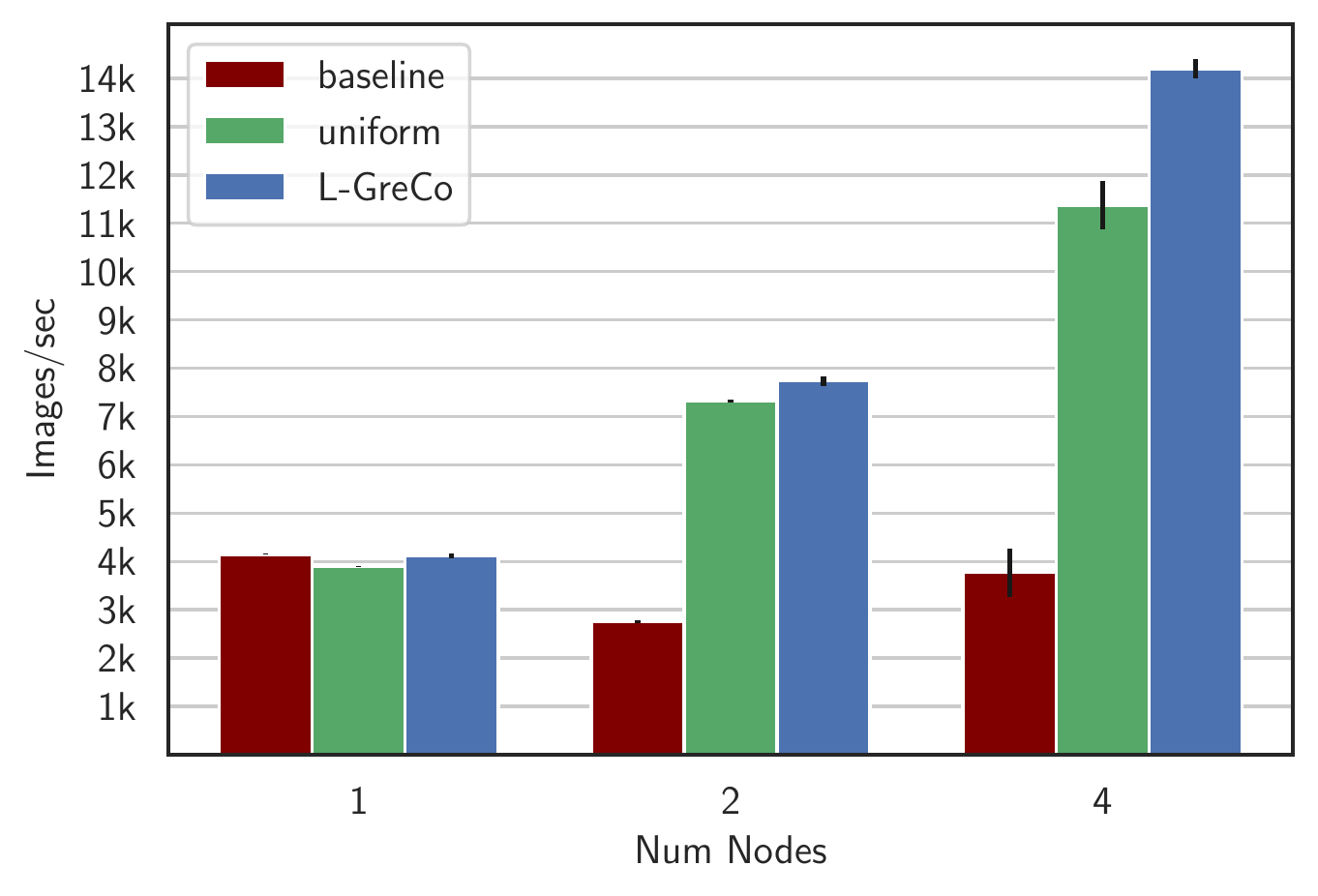}}
    \subfloat[QSGD\label{fig:rn50_multinode_qsgd}]{\includegraphics{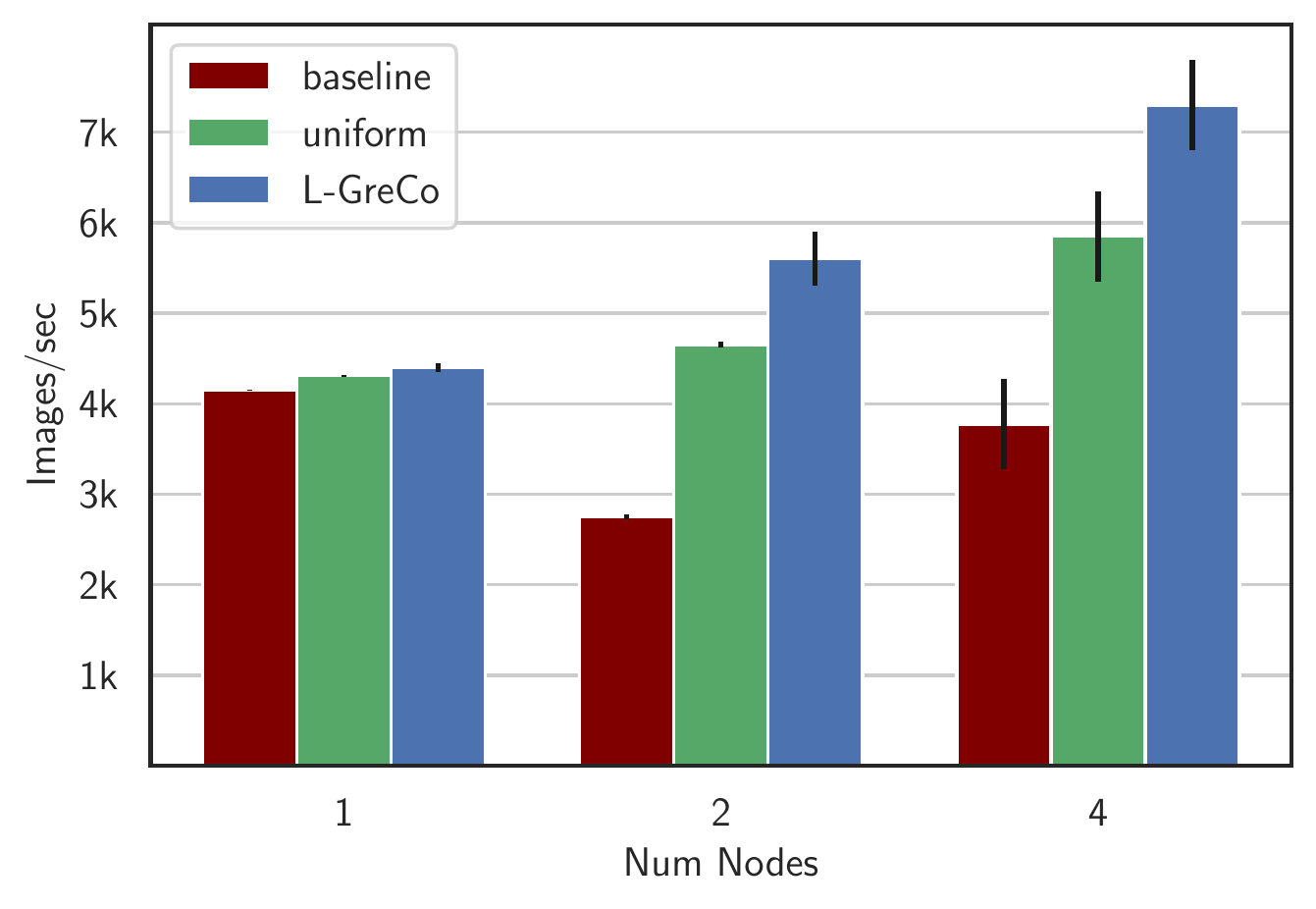}}
    \subfloat[TopK\label{fig:rn50_multinode_topk}]{\includegraphics{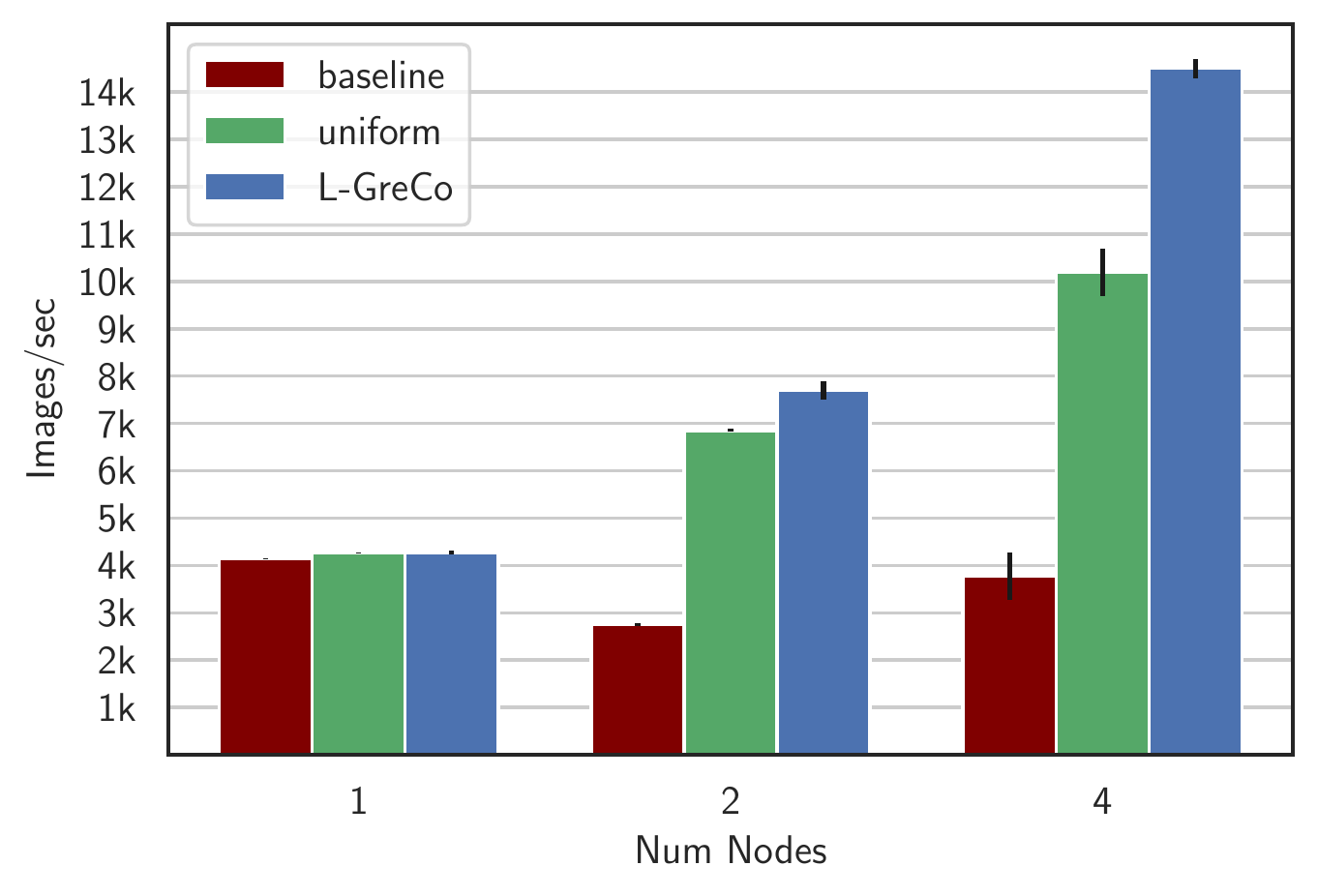}}
    \caption{{\small Throughput for ResNet50/ImageNet. Multi node, each node has 4 RTX3090 GPUs.}}
    \label{fig:rn50_multi}
 \end{figure*}

\begin{figure*}[h]
    \centering
    \setkeys{Gin}{width=0.33\linewidth}    
    \subfloat[PowerSGD\label{fig:txl_single_psgd}]{\includegraphics{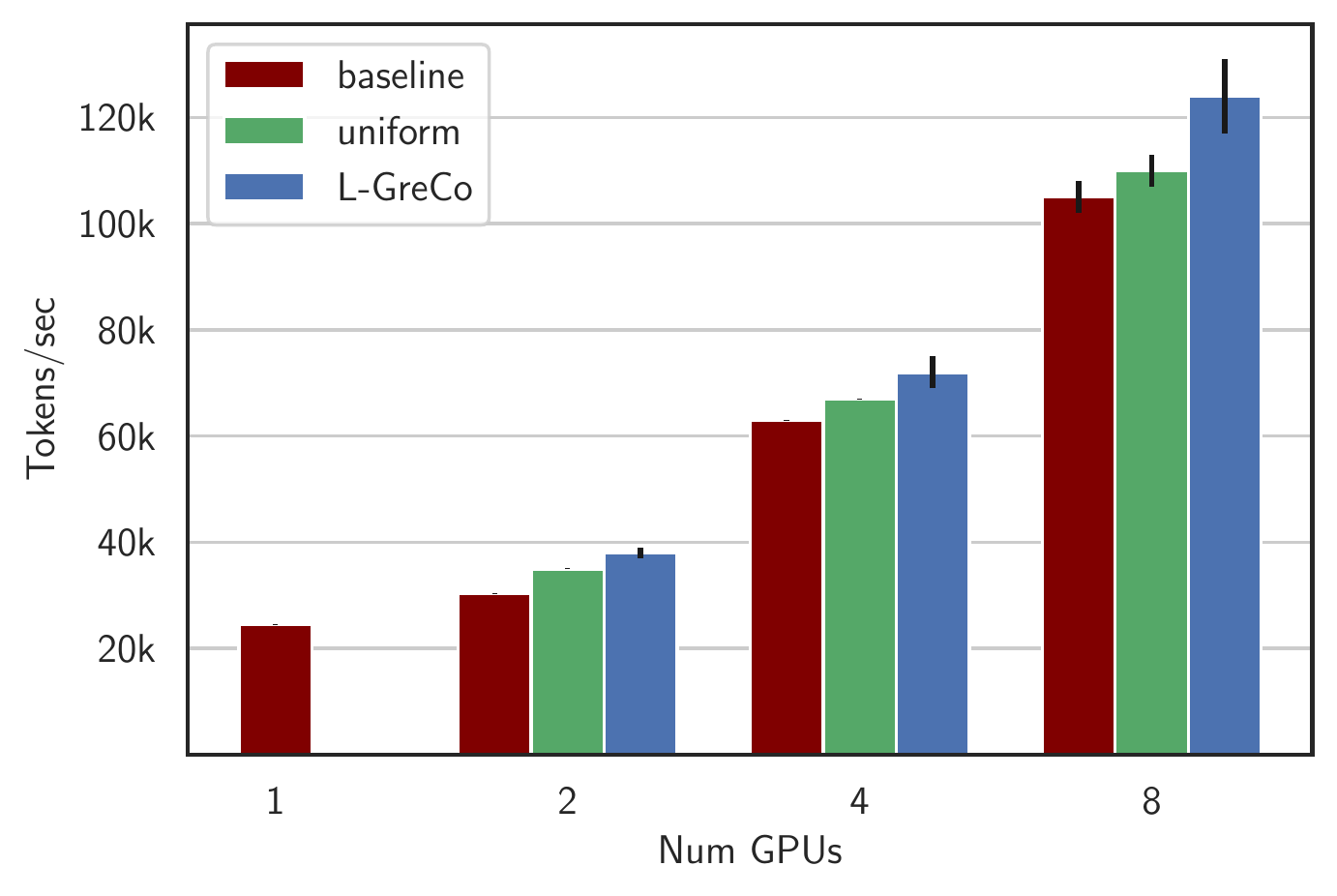}}
    \subfloat[QSGD\label{fig:throughput-txl}]{\includegraphics{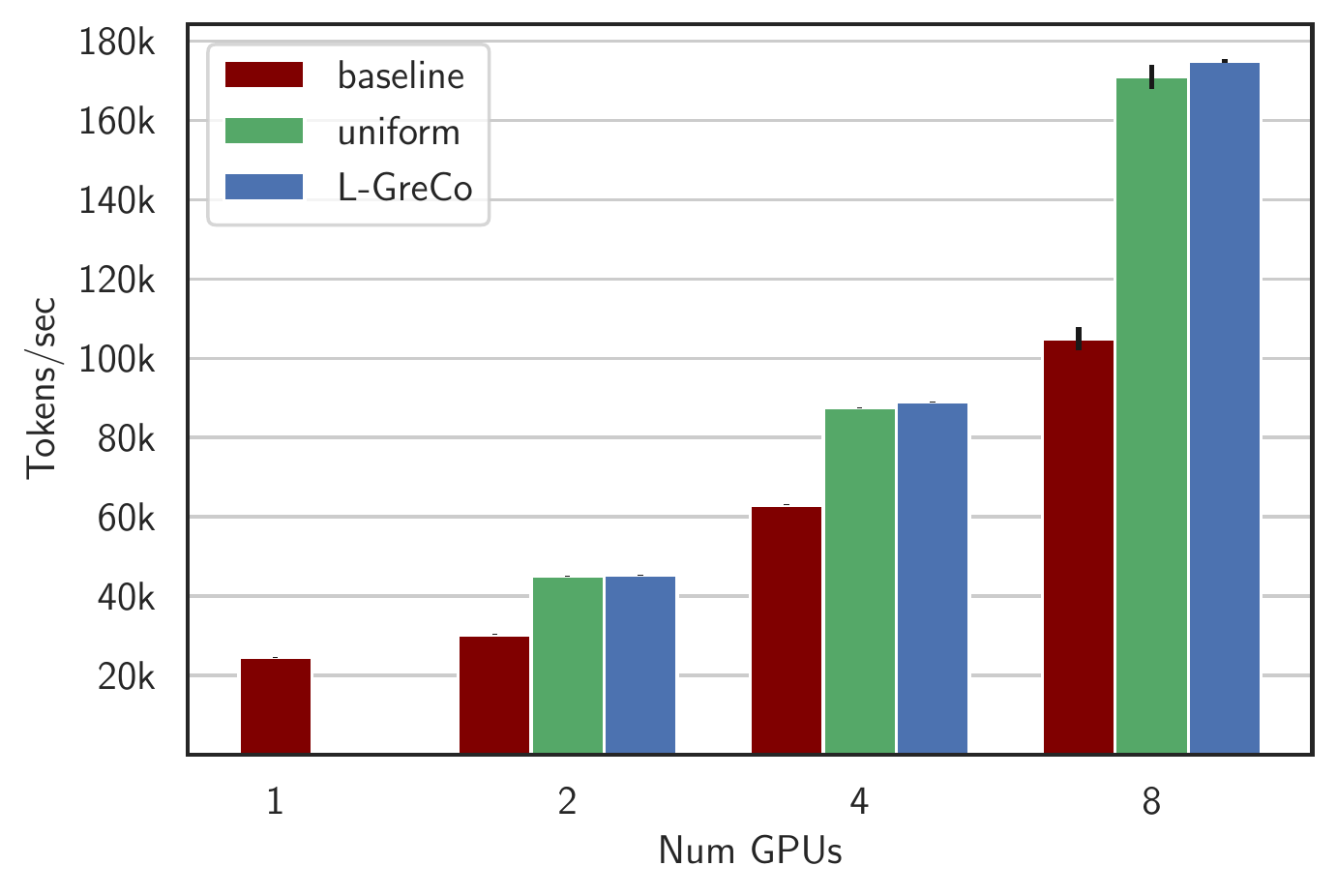}}
    \subfloat[TopK\label{fig:throughput-bert}]{\includegraphics{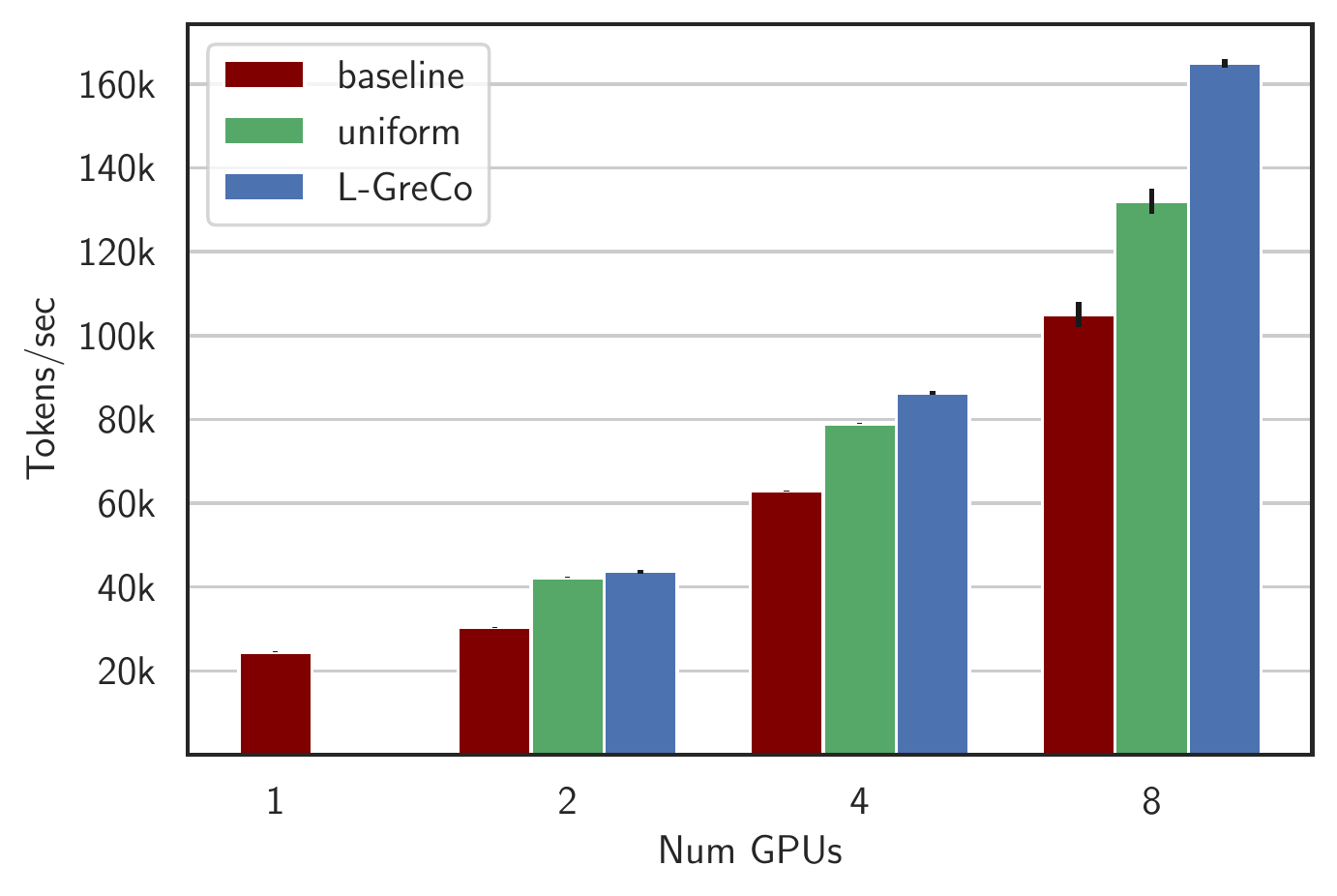}}
    \caption{{\small Throughput for Transformer-XL (TXL) on WikiText-103. Single node, 8 RTX3090 GPUs.}}
    \label{fig:txl_single}
 \end{figure*}

\begin{figure*}[h]
    \centering
    \setkeys{Gin}{width=0.33\linewidth}    
    \subfloat[PowerSGD\label{fig:rn50_single_psgd}]{\includegraphics{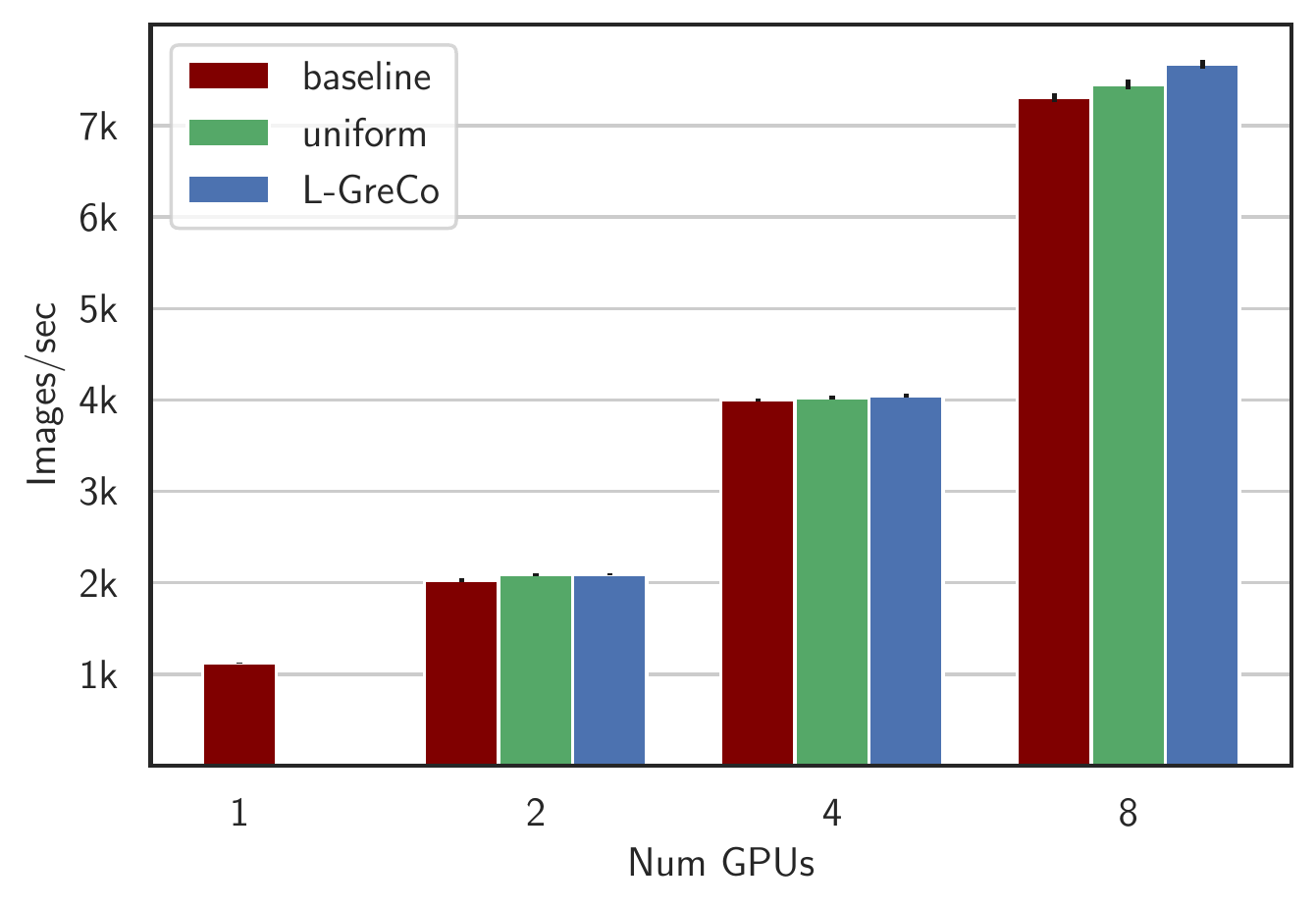}}
    \subfloat[QSGD\label{fig:rn50_single_qsgd}]{\includegraphics{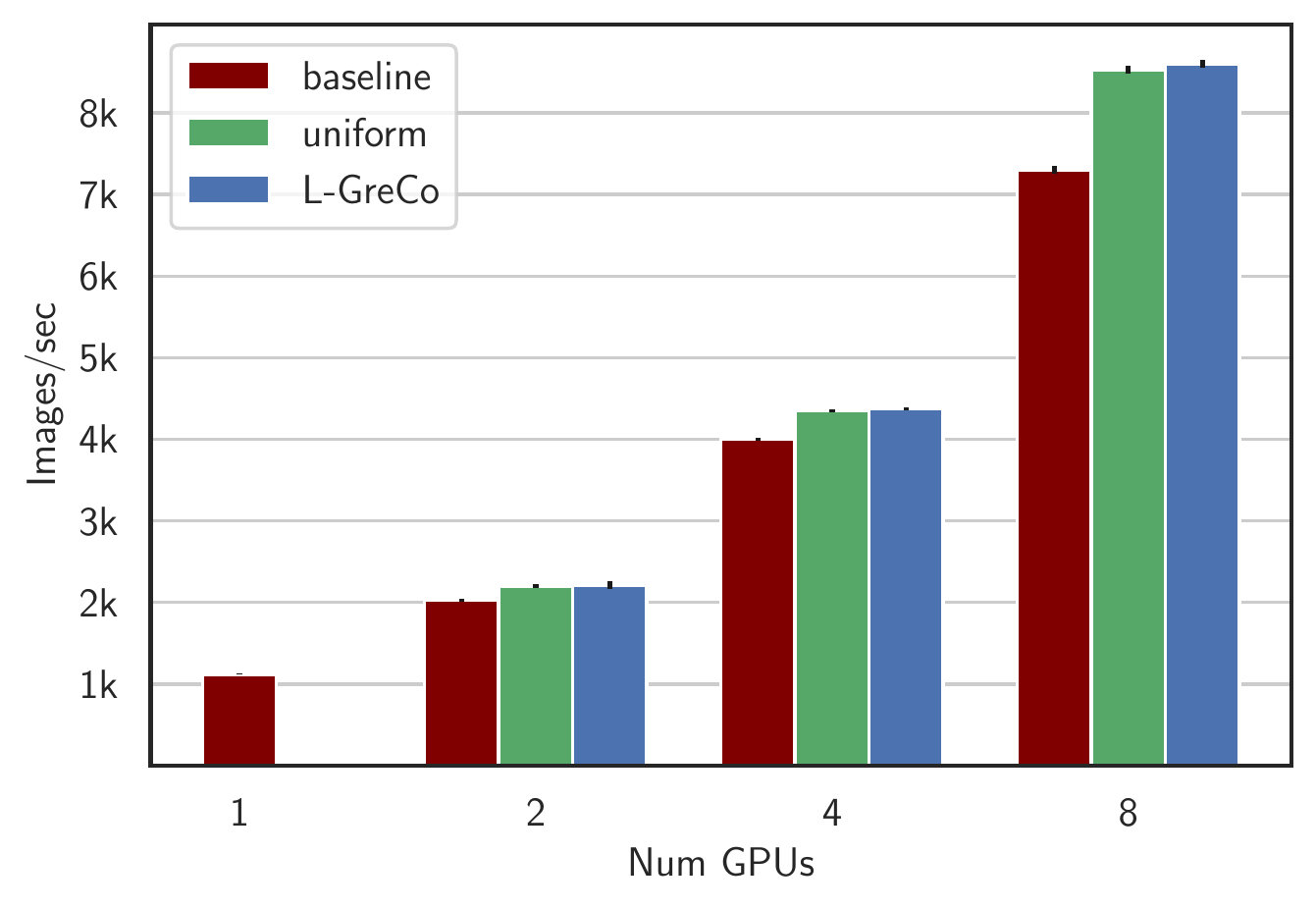}}
    \subfloat[TopK\label{fig:rn50_single_topk}]{\includegraphics{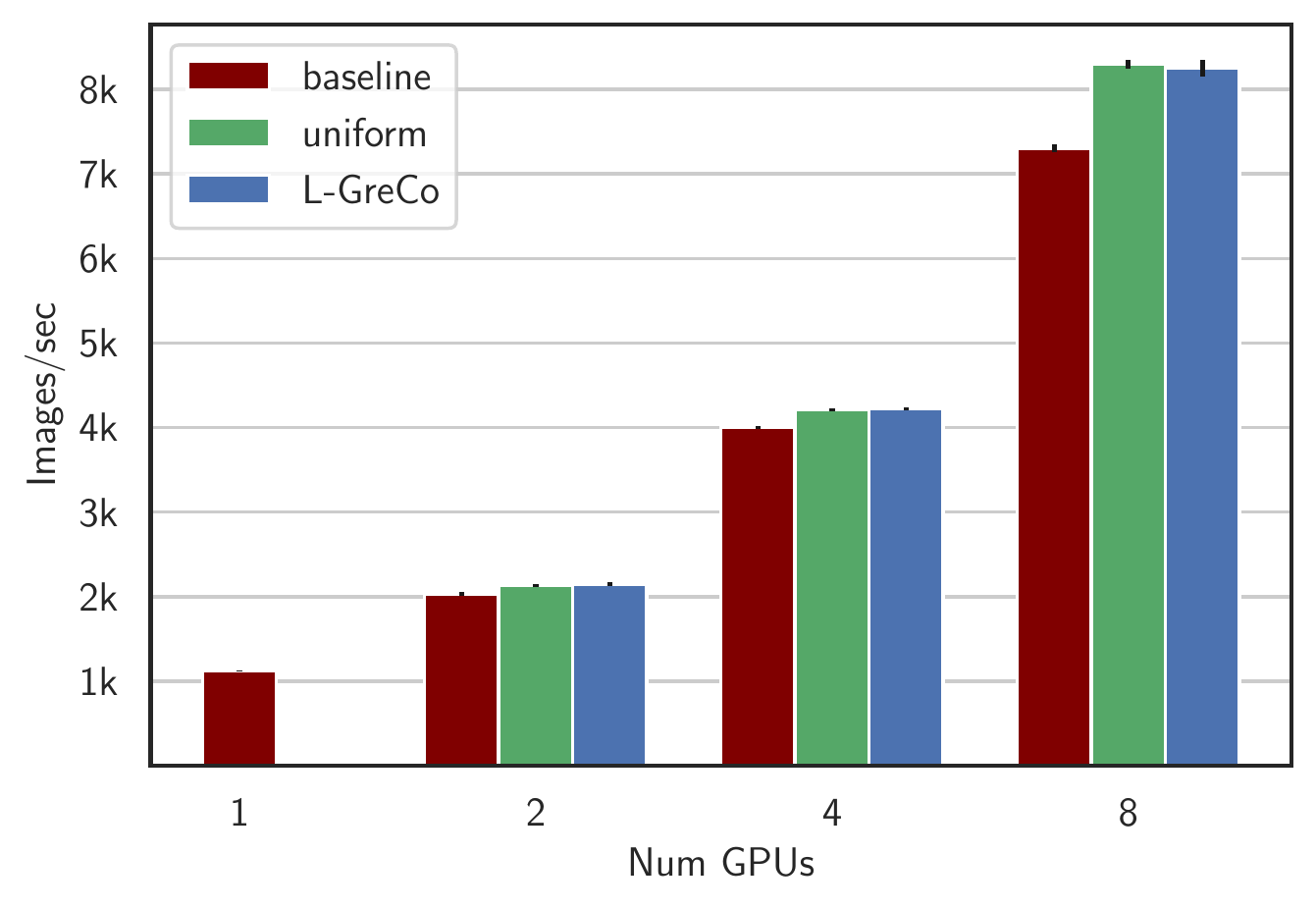}}
    \caption{{\small Throughput for ResNet50/ImageNet. Single node, RTX3090 GPUs.}}
    \label{fig:rn50_single}
 \end{figure*}

\begin{figure}[t]
 \centering
 \includegraphics[width=0.49\textwidth]{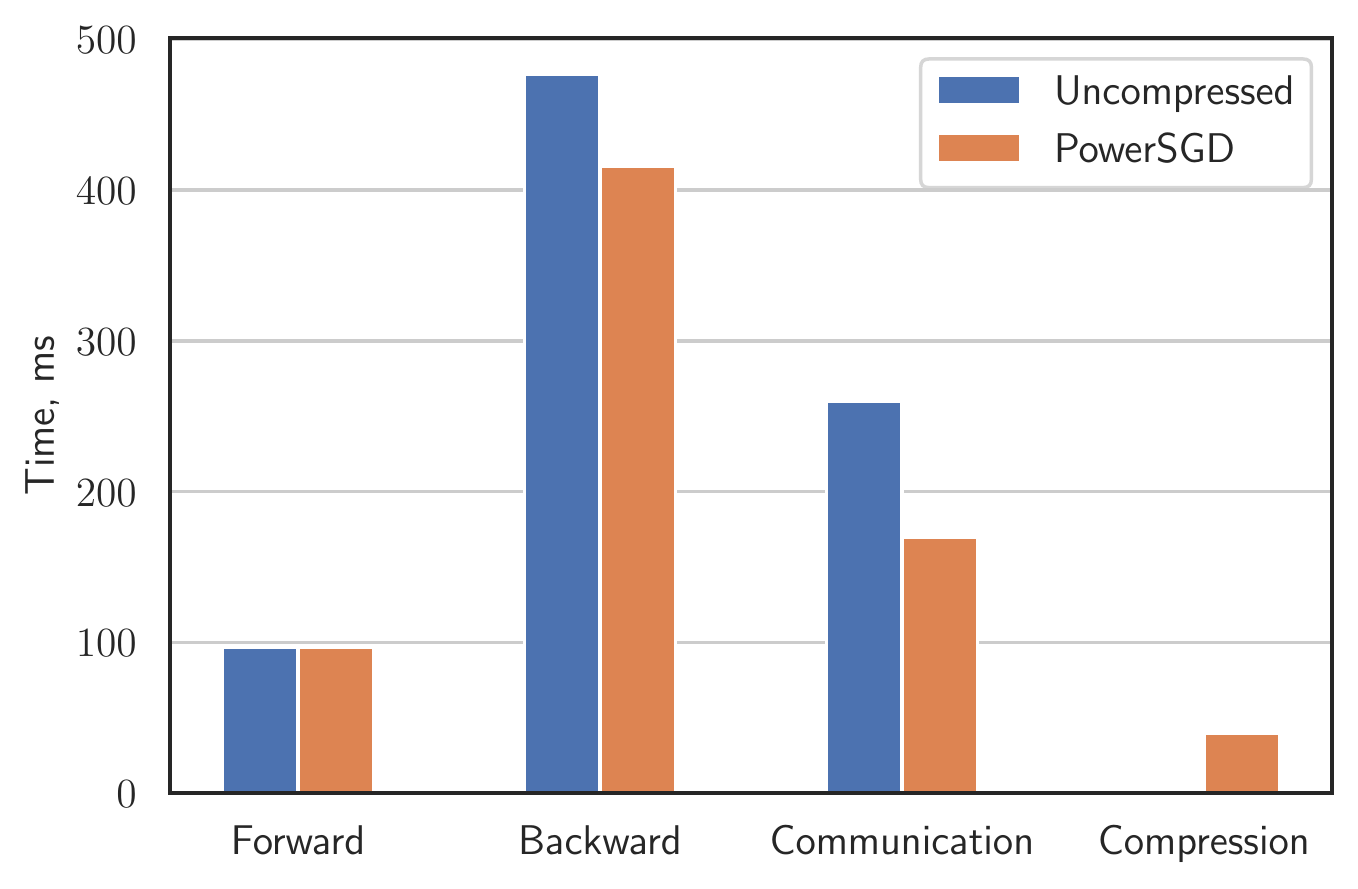}
 \caption{Profiling of the training without compression vs PowerSGD compression, rank 32. Transformer-XL model on WikiText-103 dataset. Single node, RTX3090  GPUs.}
 \label{fig:profiling}
\end{figure}

\paragraph{Profiling.}
In order to explore the compression overhead we run the profiling the training. The result is presented in Figure.~\ref{fig:profiling}. There we compare operation timings for the original(uncompressed) training and training where the gradients are compressed with PowerSGD, rank 32. We can see that relatively expensive compression (PowerSGD is more time-consuming than QSGD and optimized TopK) takes less than 10\% of the step time.

\paragraph{Speedup results.}
For end-to-end training speedup improvements, we used standard uncompressed training hyperparameters for ResNets and Transformers. We consider \emph{weak scaling}, i.e. increase the global batch size while increasing the node count. (Performance improvements are higher for strong scaling.) 
We begin by examining training throughput results for \emph{multi-node} experiments in Figures~\ref{fig:txl_multi} and~\ref{fig:rn50_multi}, executed in the cloud environment. 
This setting encounters a bandwidth bottleneck even at a lower node count, which is apparent given the poor performance of the uncompressed baseline for both models. 
Tuned uniform compression removes this bottleneck to a significant extent: for instance, uniform PowerSGD/ResNet50 can reach 75\% of ideal scaling on 4 nodes. 

It is therefore fairly surprising that automatic non-uniform compression can still provide  significant improvements in this setting: 
relative to uniform compression,  \texttt{\projectName}  gives up to 1.45x throughput improvement for ResNet50, and up to 2.4x speedup on Transformer-XL. 
This suggests that non-uniform compression can be an effective strategy in this scenario, especially for layer-heterogeneous models such as Transformers.

We next examine results for single-node scaling from 1 to 8 GPUs, presented in  Figures~\ref{fig:txl_single} and~\ref{fig:rn50_single}. 
This is a more challenging scenario for communication compression since bandwidth is less of a bottleneck in a single-node setting. 
We begin by examining the results for the Transformer-XL model. For PowerSGD and TopK, \texttt{\projectName} leads to gains up to  25\% \emph{end-to-end} speedup compared to uniform, with negligible accuracy difference. 
For QSGD, the search space is very limited: uniform already uses 4 bits, and provides very good scaling. Our adaptive method still provides 2\% speedup compared to our well-tuned uniform compression, and 50\% speedup compared to non-compressed training, reaching $\geq 90\%$ of ideal scaling.
Here, we observe that, for ResNet50, PowerSGD and QSGD are able to recover ResNet50 accuracy with uniform parameters that are close to theoretically-optimal ones (e.g. rank-4 for PowerSGD, whereas the minimum is rank-1), leaving little room for improvement given  adaptive mechanisms. Nevertheless, \texttt{\projectName} provides consistent improvements in the case of the heterogeneous Transformer-XL model (Figure~\ref{fig:txl_single}). 


 Overall, we note that \texttt{\projectName} provides statistically-significant performance improvements over static uniform compression (especially given heterogeneous models) when applied to all considered compression methods, with negligible impact on accuracy.

\subsection{Comparison with other adaptive methods}
\label{sec:accordion}
So far, we have used uniform compression as our baseline. We now compare \texttt{\projectName} with  prior works on adaptively choosing compression parameters. We consider the  \texttt{Accordion}~\cite{agarwal2021adaptive} and \texttt{CGX}~\cite{CGX2022} approaches, as they are the closest in terms of scope, and the most general in terms of applicable compression methods. We perform our comparison on the Transformer-XL model on WikiText-103, as 1) it is a model that is sensitive to gradient compression, 2) has high heterogeneity of layers, and 3) suffers from bandwidth bottlenecks (see Figure~\ref{fig:txl_multi}). 
PowerSGD was chosen as a compression method, as \texttt{Accordion} specifically optimizes for it, and this approach can lead to the highest practical compression gains. We run the experiments in two distributed settings: single node with 8 GPUs and multi-node which includes 4 machines with 4 GPUs each. We compare compression and throughput (processed  samples per second) for each method. 

In the experiments in this section we deviate from our approach to the choice of search range of the compression parameters for the \texttt{\projectName} algorithm, described in Section~\ref{sec:exp_setup}. Here, we aim to get the most speedup without losing the final model accuracy. Therefore, for each algorithm we consider in this section we tune the compression parameters for the chosen compression method and training task (without changing the training hyperparameters, e.g. learning rate, weight decay, etc.) so that we get the best timing results with the final model accuracy within MLPerf allowed bounds. The best range of parameters for \texttt{\projectName} turns out to be ranks $[8, 64]$ (see Table~\ref{table:throughput_other_works}) with default rank 32.

\begin{table}[h]
\centering
\caption{Comparison of \texttt{\projectName} with other adaptive algorithms on  Transformer-XL using PowerSGD.}
\label{table:throughput_other_works}
{\footnotesize
\begin{tabular}{|M{15mm}|M{10mm}|M{10mm}|M{12mm}|M{12mm}|}
\hline
Adaptive algorithm & Param. Range & Ratio &  Single node Tokens/s & Multi-node Tokens/s \\
\hline
Uniform & 32 & 14.1 &  110k & 72k \\
\hline
\projectName & 8 - 64 & 23.5 & \textbf{144k} & {150k} \\
\hline
Accordion & 16, 64 & 23.9 & 114k & 107k \\
\hline
CGX, kmeans & 8 - 64 & 21.6 & 124k & 112k \\
\hline
\projectName\ + Accordion & 8 - 128 & 36.9 & 138k & \textbf{176k} \\
\hline
\end{tabular}
}
\end{table}

\begin{figure}[t]
 \centering
 \includegraphics[width=0.49\textwidth]{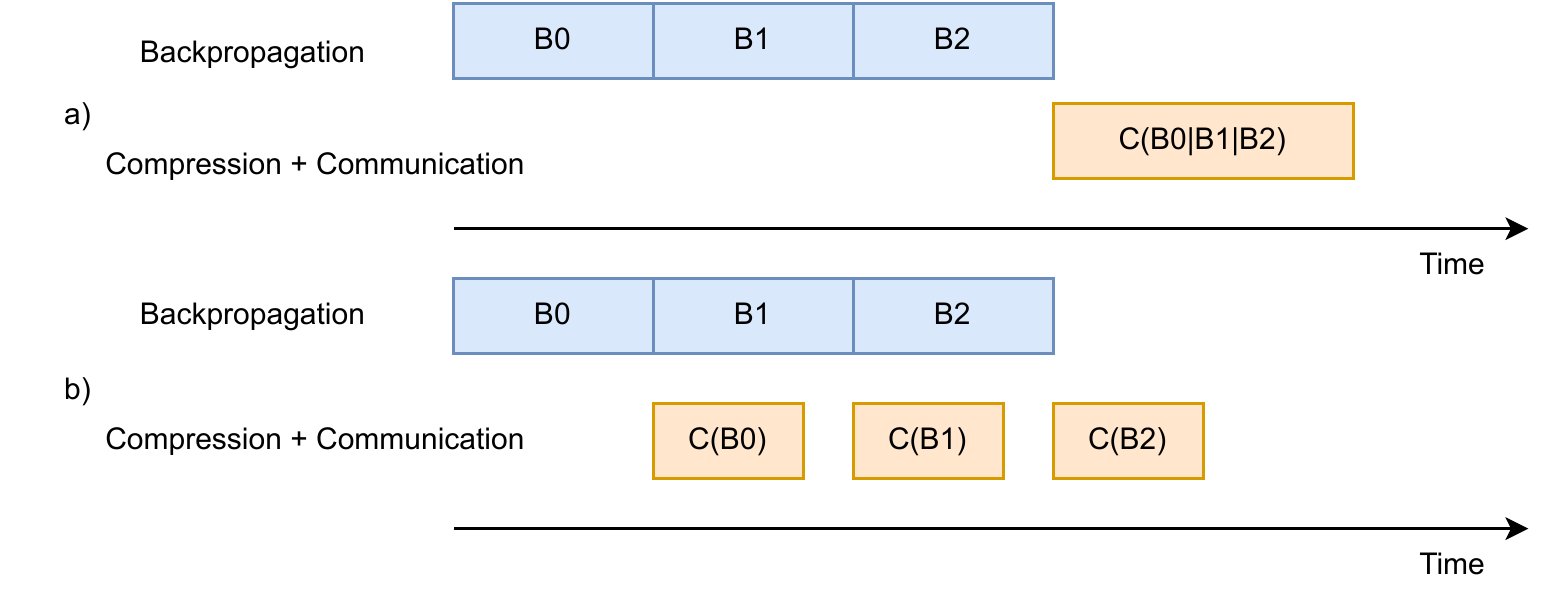}
 \caption{Scheme of communication in case of (a) Global TopK compression and (b) standard bucket-wise communication. Global TopK has to collect entire model gradients in buckets before it compresses and synchronizes them, whereas standard approach allows to overlap compression and communication with backpropagation.}
 \label{fig:global_topk}
\end{figure}

\paragraph{Global TopK Comparison.}
The optimization problem - minimizing of gradient error magnitude given the desired compression ratio - can be resolved by global topK in case of gradient sparcification. Indeed, global topK is applied to the gradients of the entire model so it selects the elements with the highest magnitudes, minimizing the total error. However, this method has several drawbacks. First, the global topK requires proper fine-tuning and search for hyperparameters in order to converge when low densities are used. \texttt{\projectName} in its turn does not try to minimize the global compression error -- it tries to match it to the compression error of the uniform \textit{layer-wise} compression that recovers the accuracy. Also, \texttt{\projectName} has lowest possible compression parameter. It means that each layer has its contribution in gradient synchronization. Which might not be a case for the global topK - at high densities some layers could be left without updates for several optimization steps. This has a bad impact on the final model quality. The second disadvantage of the global topK is an actual speedup it might provide. As we discussed in the Section~\ref{sec:comm_details}, in modern DataParallel frameworks the gradients are synchronized in parallel with computation in the sake of the efficiency - one can hide the communication costs behind the computation. However, in case of the global topK, as long as one has to wait till all the gradients are ready, then perform compression and communication. It might turn out that in this case we have more non-overlapped communication than in the layer-wise compression or even in uncompressed case (see Figure~\ref{fig:global_topk}). 

In order to confirm our doubts about the global topK we have implemented the algorithm using \texttt{torch.distributed} hooks and ran the RN18/CIFAR100 training. We found out that the global topK is 10\% slower than \texttt{\projectName} when applied with the similar global density in this case - 0.25\%.

\paragraph{Accordion Comparison.}
\texttt{Accordion} adapts compression by detecting critical regimes during training. 
It accepts two possible compression modes (corresponding to low and high compression), and has a threshold error parameter $\eta$. It collects the gradients, and periodically decides the parameter to use based on gradient information for each layer. 
We have implemented Accordion using the \texttt{torch.distributed} hook, used for PowerSGD. For the parameter $\eta$, we chose the value of $0.5$ suggested by the authors and tried to hand-tune the best pair of low and high compression parameters for each model, with which training converges to an accuracy that is within MLPerf bounds. We ran this algorithm on Transformer-XL/Wikitext-103, and found that the best pair of parameters (in terms of training time without losing accuracy) are high compression rank 16, and low compression rank 64. 

In Figure~\ref{fig:accordion}, one can see the dynamics of the average compression ratio over the training of \texttt{Accordion} and \texttt{\projectName}. We notice that \texttt{Accordion} chooses a low compression rank for almost all layers during the first period of training and a high compression rank for the rest of the training time, leading to completely bimodal uniform compression. This suggests that Accordion may not really exploit the heterogeneous nature of  DNN models. Therefore, the optimizations of \texttt{Accordion} and \texttt{\projectName}, respectively, could be seen as \emph{orthogonal}:  \texttt{Accordion} focuses on varying the amount of \emph{average compression} during training, whereas \texttt{\projectName} finds an optimal way of reaching this average level by setting layer-wise targets. 

With this in mind, we combined these two algorithms, as follows. 
We first executed \texttt{Accordion} to get the suggested parameters for each layer and used these parameters as the default set of parameters in \texttt{\projectName} algorithm. The default set of parameters is used to define the maximal error of the DP algorithm (see line 2 in Algorithm~\ref{algo:dp}). In this approach, \texttt{Accordion} determines the model sensitivity to gradient compression at different points in training, while \texttt{\projectName} finds the best mapping of compression parameters per layer. In Figure~\ref{fig:accordion}, we see that the resulting combination (\texttt{\projectName} with range $[8, 128]$  and Accordion with high=16, low=64) provides superior results in terms of compression ratio, without sacrificing accuracy.


We also compare the performance of the two algorithms in isolation (see Table~\ref{table:throughput_other_works}). 
We observe that, despite the fact that the theoretical compression ratio suggested by \texttt{Accordion} is essentially the same as that of \texttt{\projectName}, the \texttt{Accordion} throughput is less by around 30\%. 
This is explained by the fact that \texttt{\projectName} compressed the last transmitted layers (buckets) to higher levels, leading to significantly-improved total transmission time. 
Specifically, in Figure~\ref{fig:accordion_buckets}, we observe that \texttt{\projectName} transmits twice fewer elements in the last bucket relative to \texttt{Accordion}. 
Moreover, combining \texttt{\projectName} with \texttt{Accordion} improves the compression ratio by 50\%, and training time by up to 66\% compared to \texttt{Accordion}.

\paragraph{Rethink-GS comparison.}
\label{app:rething_gs}
\citet{sahu2021rethinkinggs} suggest a sparcification method which is technically adaptive - hard-threshold sparsification changes number of transmitted elements based on the gradient distribution.
We have run the \texttt{\projectName} training of ResNet18/CIFAR-100 in the setup described in the paper. We used 1\% density as a default parameter for \texttt{\projectName} and the search range was $[0.1\%, 10\%]$. For \texttt{Rethink-GS} we used parameter $\lambda = 4.72 \times 10^{-3}$. 

As a result we saw that \texttt{\projectName} improves \texttt{Rethink-GS} algorithm by 17\% in terms of compression ratio ($6.7\times$ vs $5.7\times$) while also improving the final accuracy - 71.7\% vs 71.4\% (the numbers differ from the ones we show in Table.~\ref{table:results_ic} as here we used the setup from \cite{sahu2021rethinkinggs}). We note that our framework did not require any hyperparameter tuning at all for this experiment, whereas Rethink-GS requires careful tuning of the hard-threshold $\lambda$ parameter.

\paragraph{CGX Comparison.}
The adaptive compression of \texttt{CGX} is based on \texttt{kmeans} and maps layers into a 2-dimensional space (layer size vs. $L_2$-error). 
The algorithm clusters layers into several groups and assigns predefined compression parameters to the layers in the groups. We have implemented this logic with PowerSGD compression. We used rank 32 as default, and the best (in terms of compression) range was from 8 to 64, using 6 layer clusters.
The results are shown in the Table~\ref{table:throughput_other_works}. In short, \texttt{\projectName} improves upon the kmeans approach by up to 33\%. In  Figure~\ref{fig:accordion_buckets}, we observe that \texttt{\projectName} picks parameters such that the largest and the last bucket are compressed the most, whereas the \texttt{kmeans} algorithm chooses worse compression parameters for those layers in some iterations.

Overall, \texttt{\projectName} improved practical  compression relative to prior techniques. Of note, the highest compression ratio is achieved by the hybrid \texttt{Accordion} + \texttt{\projectName} method, which leverages layer-wise insights in terms of both sensitivity and training dynamics. 

\begin{figure*}[t]
     \centering
     \setkeys{Gin}{width=0.49\linewidth}
     \subfloat[Metrics correlation\label{fig:txl_metrics_corr}]{\includegraphics{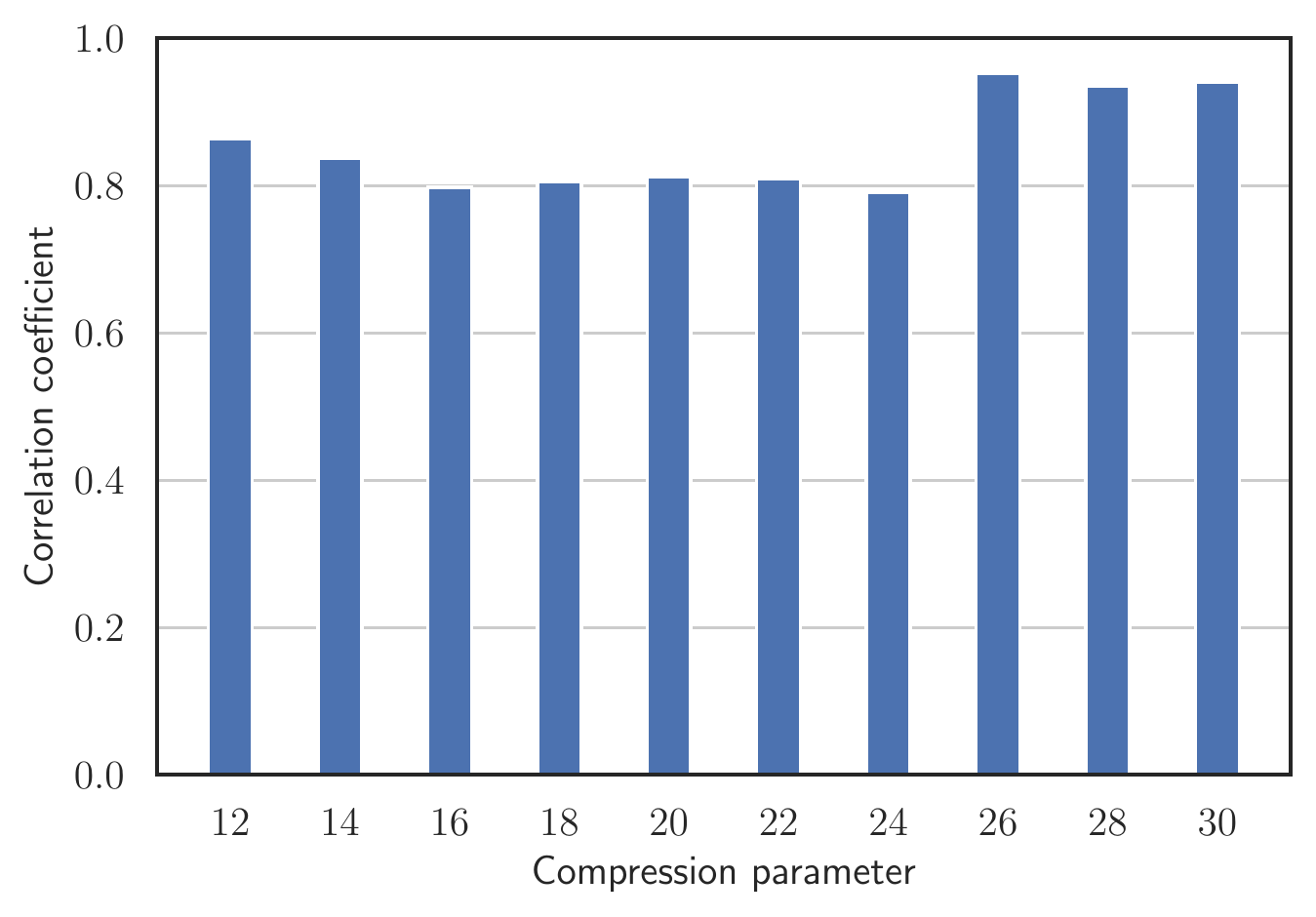}}
    \subfloat[Coefficients\label{fig:txl_timing_coefs}]{\includegraphics{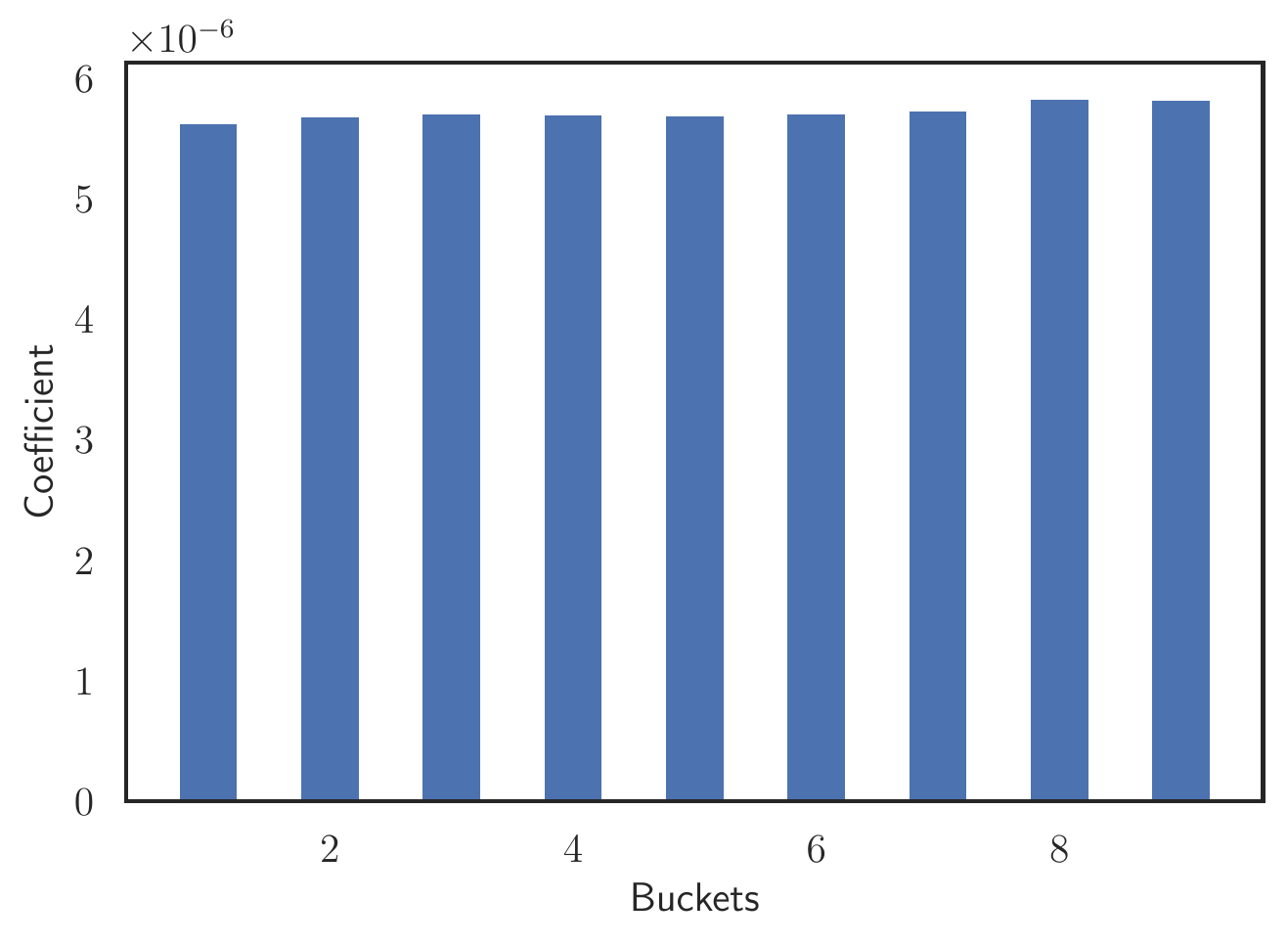}}
     \caption{Correlation coefficients between metric values (a)  for PowerSGD using loss-based and error-magnitude approaches as the sensitivity metrics. Timing coefficients per bucket (b) for timing-based approach. Transformer-XL/WikiText-103 model training with PowerSGD method. }
\end{figure*}

\subsection{Loss-based accuracy metric}
\label{sec:experiments_metric}
As we discussed in the~\ref{sec:metric} it is possible to use different metric for measuring the layer sensitivity to compression. The one we used in the all experiments is error magnitude. The other possible one is loss-based. in which we collect the model loss differences between uncompressed training and training with the gradient compression of certain layers while other layers stay intact and use the loss difference as a sensitivity metric.

Comparing these two approaches, we evaluate the correlation coefficients of the metrics these two approaches provide. We can see in the figure~\ref{fig:txl_metrics_corr}, the metric values from two approaches have a high correlation. Moreover, we found out in our experiments that the compression parameters as well as the total compression ratio resulted for both metrics are very close to each other. Hence, given the fact that collecting loss-based metrics requires additional off-training runs, our usage of error-based metric is justified.

\subsection{Timing-based optimization}
\label{sec:time_based}

Considering the fact that communication buckets have different impact on training performance, one may notice that compression ratio does not always give an expected speed improvement. Having that in mind we modified \texttt{\projectName} algorithm so that it optimizes the expected time of the communication rather than compression ratio. As we explained in Section~\ref{sec:time_optimization}, in this approach we measured the communication time varying compression ratio for each communicated bucket and trained linear regression model, running the training for 50 steps with 5 steps warmup for each set of compression parameters. With the model we get the weights of the buckets that correspond to a relation between number of communicated bits per bucket and the communication time. Then we use those coefficients in the metric we try to minimize in the Algorithm~\ref{algo:dp}.

We have run the algorithm on our workhorse benchmark - PowerSGD in Transformer-XL/WikiText-103 training. The linear model built on the timing data we collected (5000 samples - sets of compression ratios per bucket) has a score close to 1 meaning that we managed to predict the communication time using communicated bucket sizes almost perfectly. We found out that the per-bucket coefficients from the linear model are close to each other (see Figure.~\ref{fig:txl_timing_coefs}). It means that each bucket's impact on communication time is proportional to the bucket size. We noticed that the parameters we get with the modified algorithm are very close to the parameters from the original \texttt{\projectName} algorithm. Given that we figure that in the case of Transformer-XL/WikiText-103 the original algorithm gives the most optimal in terms of timing parameters.

\section{Conclusion}

We proposed \texttt{\projectName}, an adaptive gradient compression algorithm which automatically identifies optimal layer-wise compression parameters, given a fixed error constraint. 
The \texttt{\projectName} algorithm finds the mapping of compression parameters such that 1) the total $L_2$ compression error matches a target known to recover accuracy, and 2) the total compressed size is minimal for this target. 
%

Our experimental validation across all families of gradient compression methods show that training with the layer-wise parameters suggested by \texttt{\projectName} recovers the baseline accuracy while the gradient compression ratio is substantially increased. \texttt{\projectName} improves training performance up to $2.5\times$, saving up to $5.2$x communication compared to vanilla  compression, and up to $122$x relative to uncompressed training. Overall, our work provides a new approach for improving existing gradient compression methods, at almost zero cost in terms of time and accuracy loss.


Our approach suggests that per-layer gradient compression may be an interesting direction for further work in this area. Specifically, our method can be augmented with actual timing information regarding the layer gradient transmission cost, which could lead to further practical gains. Another possible extension would be considering \emph{hybrid} strategies, which allow for combining different compression techniques (e.g. sparsification and low-rank) inside the same model. 

We proposed \texttt{\projectName}, an adaptive gradient compression algorithm which automatically identifies optimal layer-wise compression parameters, given a fixed error constraint. 
The \texttt{\projectName} algorithm finds the mapping of compression parameters such that 1) the total $L_2$ compression error matches a target known to recover accuracy, and 2) the total compressed size is minimal for this target. 

Our approach is complemented by an in-depth exploration of the ``correct'' metrics which capture the accuracy and performance impact of compression, at the per-layer level. 
Specifically, we present evidence that minimizing local, per-layer quantization errors leads to very similar results to minimizing global metrics such as output loss. 
Moreover, maximizing per-layer compression rates correlates very well to specifically minimizing total transmission time on existing data-parallel implementation. 

%

We complemented these algorithmic and analytic contributions with an extensive experimental validation, across all families of gradient compression methods show that training with the layer-wise parameters suggested by \texttt{\projectName} recovers the baseline accuracy while the gradient compression ratio is substantially increased. \texttt{\projectName} improves training performance up to $2.5\times$, saving up to $5.2$x communication compared to vanilla  compression, and up to $122$x relative to uncompressed training. Overall, our work provides a new approach for improving existing gradient compression methods, at almost zero cost in terms of time and accuracy loss.
Possible extensions of our method could consider \emph{hybrid} strategies, which allow for combining different compression techniques (e.g. sparsification and low-rank) inside the same model. 

\section*{Acknowledgments}
The authors gratefully acknowledge funding from the European Research Council (ERC) under the European Union's Horizon 2020 research and innovation programme (grant agreement No 805223 ScaleML), as well as experimental support from the IST Austria IT department, in particular Stefano Elefante, Andrei Hornoiu, and Alois Schloegl.

\clearpage
\bibliography{references}
\bibliographystyle{mlsys2023}

\clearpage
\appendix

\section{Buckets prioritization}
\begin{figure}[h]
 \centering
 \includegraphics[width=0.49\textwidth]{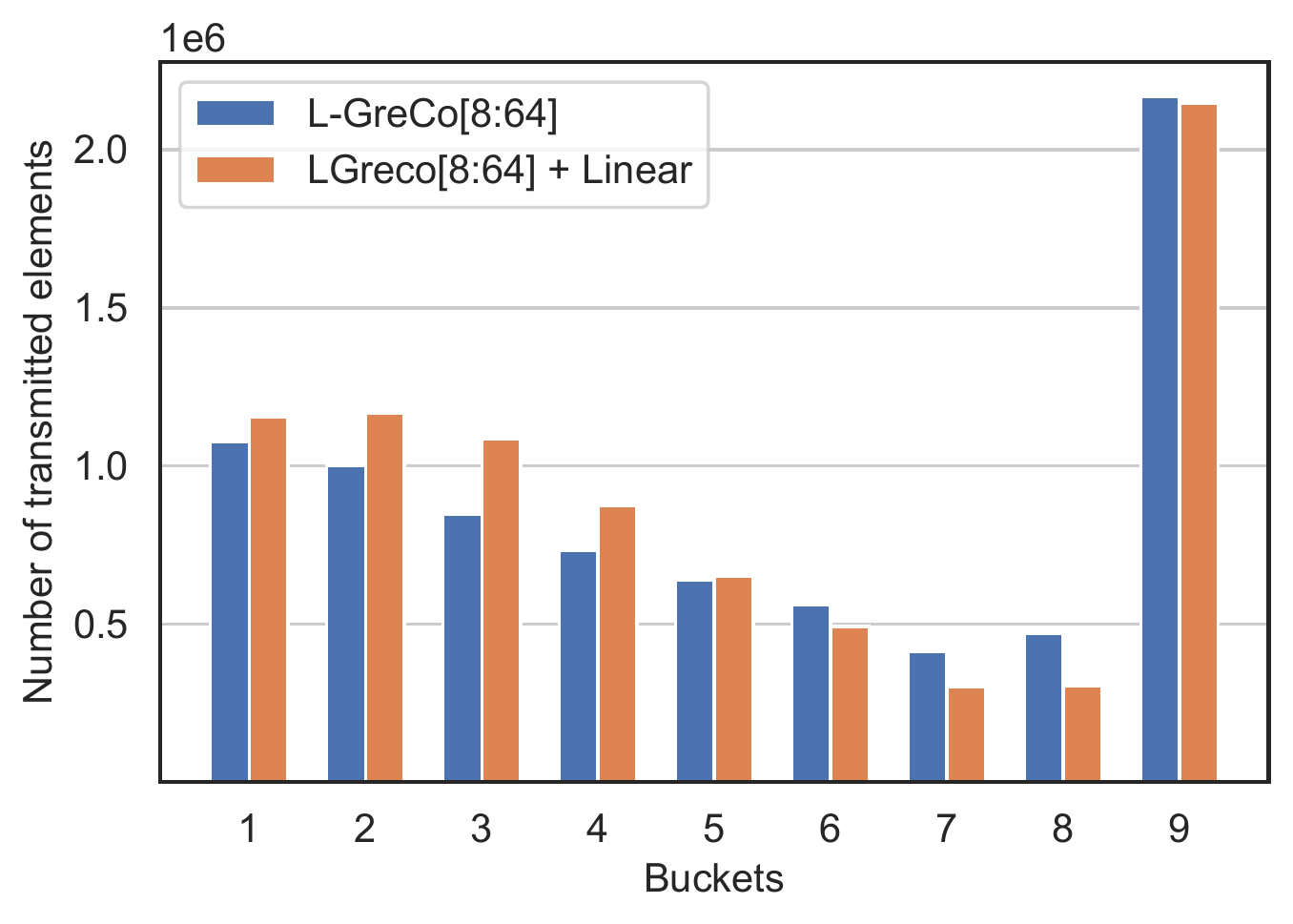}
 \caption{Communicated elements per bucket for L-GreCo and L-GreCo with linear bucket prioritizing.  Transformer-XL with PowerSGD.}
 \label{fig:bucketwise_linear}
\end{figure}

Considering the fact that communication buckets have different impact on training performance, we modified the \texttt{\projectName} algorithm so that the last buckets in transmission order, corresponding to the earlier layers, were compressed more. This compensates for the compression error caused by picking lower compression parameters for the first buckets, i.e. the last layers. In practical terms, we have added linear priorities to the layers in Algorithm~\ref{algo:dp}, multiplying the size of each layer by the index of the bucket the layer is communicated in. 
The profile of communicated elements per buckets is shown in the Figure~\ref{fig:bucketwise_linear}. We observe the linear shift of higher compression ratios towards the last buckets. However, bucket prioritization has worse performance than original \texttt{\projectName}. It means that the effect of the first big buckets transmission is higher than the effect of better compression the last buckets. 


\section{Low-rank error computation}
As discussed in Section~\ref{sec:algorithm}, one of the main steps of our algorithm is to compute the error matrix for different possible compression parameters. Table~\ref{table:overheads} suggests that this is the most time-consuming part of our framework. Specifically for the PowerSGD, we need to compute low-rank errors for a wide range of ranks. There are two possible solutions to do so.

\subsection{Singular Value Decomposition}
{The first way to compute errors is to use singular value decomposition and compute singular values for a particular layer, and calculate the approximation error for rank $r < min(m, n)$ by calculating $e_r = \sqrt{\sum_{i = r+1}^{min(m, n)} \sigma_{i}^{2}}$, which can be done efficiently for all ranks. Specifically, it is sufficient to compute squared singular values once, and then compute all the errors by a single matrix product. Thus, the bottleneck is computing singular values requiring $O(m n \cdot min(m,n))$ time and $O(n^2 + m n)$ space.}
\subsection{Power Iteration Steps}
{The second approach is to calculate the approximation error for each rank separately by doing a few power steps (without the communication parts); as \citet{vogels2019powersgd} claims, this approach converges to the SVD-suggested matrix. On the practical side, we have observed that applying only 5 power steps is enough to have a small error relative to the optimal low-rank approximation suggested by SVD. This approach needs $O(m n r)$ time and $O((m+n) \cdot r)$ space for calculating rank r approximation error and therefore $O(m n r_{max}^{2})$ to compute errors for all $r \in [r_{min}, r_{max}]$.}
\subsection{The Best of Both Worlds}

Comparing computational complexity and memory requirements of two methods suggests it is better to use the power method when the rank range is small, e.g., ResNet50 on ImageNet or ResNet18 on Cifar100, and to use the SVD method when the rank range is large, e.g., TransformerXL and TransformerLM on WIKITEXT-103.

\section{Combination of PowerSGD and \projectName}

\begin{figure}[h]
 \centering
 \includegraphics[width=0.49\textwidth]{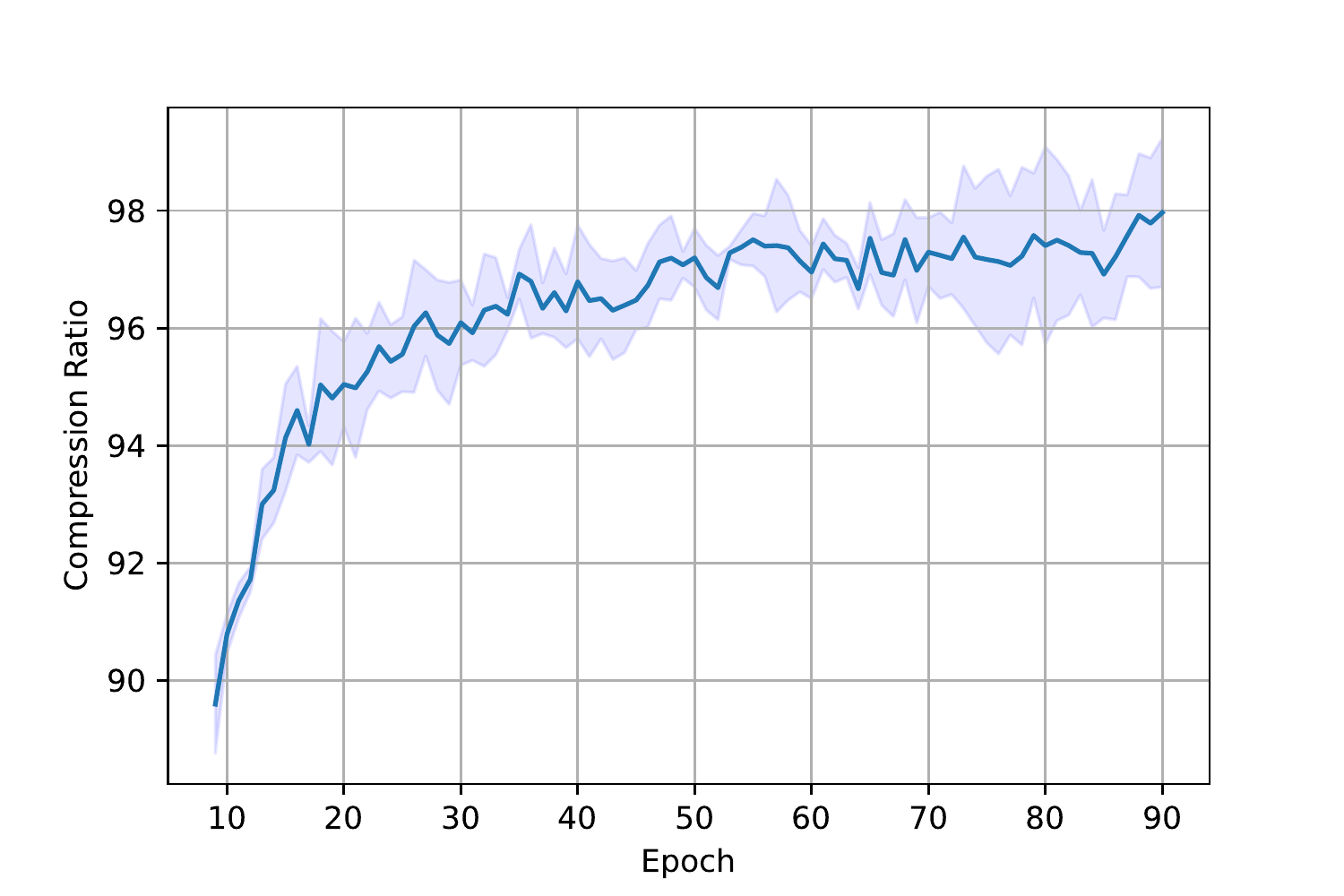}
 \caption{Compression ratio of the scheme suggested by \texttt{\projectName} during the training. ResNet50 with PowerSGD.}
 \label{fig:increasing_cr}
\end{figure}

We note that in all of our wide-range experiments, the compression ratio when \texttt{\projectName} is applied to PowerSGD generally increases during the training (see Figure~\ref{fig:increasing_cr}). This is also aligned with the intuition behind the results of \cite{Agarwal2021}. This suggests that in this scenario, \texttt{\projectName} is able to increase the compression in the less crucial learning periods, (e.g.last epochs).

\begin{table}[t]
\centering
\caption{Hyperparameters for ResNet-18/CIFAR-100}
\label{table:hyperparam_rn18}
{\footnotesize
\begin{tabular}{|M{30mm}|M{30mm}|}
\hline
Parameter & Value \\
\hline
Number of workers & 8 \\
\hline
Optimizer & SGD with momentum \\
\hline
Global batch size & 128 \\
\hline
Momentum & 0.9 \\
\hline
Base LR & 0.1 \\
\hline
LR decay & /10 at epoch 150 and 250 \\
\hline
Epochs & 200 \\
\hline
Weight decay & $10^{-4}$\\
\hline
\end{tabular}
}
\end{table}

\begin{table}[hbt!]
\centering
\caption{Hyperparameters on ResNet-50/ImageNet}
\label{table:hyperparam_rn50}
{\footnotesize
\begin{tabular}{|M{30mm}|M{30mm}|}
\hline
Parameter & Value \\
\hline
Number of workers & 8 \\
\hline
Optimizer & SGD with momentum \\
\hline
Global batch size & 2048 \\
\hline
Momentum & 0.875 \\
\hline
LR warmup & Linear for 8 epochs, starting from 0.256 \\
\hline
LR schedule & cosine \\
\hline
LR decay & /10 at epoch 150 and 250 \\
\hline
Epochs & 90 \\
\hline
Weight decay & 1/32768\\
\hline
Label smoothing & 0.1\\
\hline
\end{tabular}
}
\end{table}

\begin{table}[hbt!]
\centering
\caption{Hyperparameters on Transformer-XL/WikiText-103}
\label{table:hyperparam_txl}
{\footnotesize
\begin{tabular}{|M{30mm}|M{30mm}|}
\hline
Parameter & Value \\
\hline
Number of workers & 8 \\
\hline
Optimizer & LAMB \\
\hline
Global batch size & 256 \\
\hline
LR warmup & Linear for 1000 steps \\
\hline
LR schedule & cosine \\
\hline
Number of steps & 40k \\
\hline
Weight decay & 0.0\\
\hline
\end{tabular}
}
\end{table}

\begin{table}[t]
\centering
\caption{Hyperparameters on Transformer-LM/WikiText-103}
\label{table:hyperparam_tlm}
{\footnotesize
\begin{tabular}{|M{30mm}|M{30mm}|}
\hline
Parameter & Value \\
\hline
Number of workers & 8 \\
\hline
Optimizer & Adam \\
\hline
Adam betas & (0.9, 0.98) \\
\hline
Global batch size & 2048 \\
\hline
LR warmup & Linear for 4000 steps starting from $10^{-7}$ \\
\hline
LR schedule & inverse sqrt \\
\hline
Number of steps & 50k \\
\hline
Weight decay & 0.01 \\
\hline
\end{tabular}
}
\end{table}

\newpage
\section{Detailed experimental settings}
For all the experiments we used the standard hyperparameters, datasets and data preprocessing. The detailed hyperparameters are shown in the tables~\ref{table:hyperparam_rn18},~\ref{table:hyperparam_rn50},~\ref{table:hyperparam_txl}, and ~\ref{table:hyperparam_tlm}.

For preprocessing the images of CIFAR-100 datasets, we follow the standard data augmentation and normalization routines. Random cropping and horizontal random flipping were applied for data augmentation. We also normalized each color with the following mean and standard deviation values for each channel: (0.4914, 0.4822, 0.4465) and (0.2023, 0.1994, 0.2010).

ResNet50 model uses the following data augmentation. We perform rand resized crop to $224\times224$, scale from 8\% to 100\% and do a random horizontal flip. Also, we do normalization with means (0.485, 0.456, 0.406) and standard deviations (0.229, 0.224, 0.225).

For wikitext-103 preprocessing we used the standard preprocessing tools and tokenizers provided by Nvidia Examples~\cite{DeepLearningExamples} and FairSeq library~\cite{ott2019fairseq}.



\end{document}